\documentclass{article}

\usepackage[numbers]{natbib}
\usepackage{wrapfig}
\usepackage{bbm}
\usepackage{algorithm}
\usepackage{algpseudocode}
\usepackage[dvipsnames]{xcolor}
\usepackage{multirow}
\usepackage{subcaption}

\usepackage[preprint,nonatbib]{neurips_2025}

\usepackage[utf8]{inputenc} 
\usepackage[T1]{fontenc}    
\usepackage{hyperref}       
\usepackage{url}            
\usepackage{booktabs}       
\usepackage{graphicx}      
\usepackage{amsfonts}       
\usepackage{nicefrac}       
\usepackage{microtype}      
\usepackage{xcolor}         

\title{What is the role of memorization in Continual Learning?}

\author{%
  Jędrzej Kozal\thanks{Department of Systems and Computer Networks, Wrocław University of Science and Technology} \\
  Wrocław Tech\\
  Wrocław, Poland \\
  \texttt{jedrzej.kozal@pwr.edu.pl} \\
  \And
  Jan Wasilewski\thanks{Chester F. Carlson Center for Imaging Science, Rochester Institute of Technology} \\
  RIT \\
  Rochester NY, USA \\
  \texttt{jw7630@g.rit.edu} \\
  \And
  Alif Ashrafee\footnotemark[2] \\
  RIT \\
  Rochester NY, USA \\
  \texttt{aa5264@rit.edu} \\
  \AND
  Bartosz Krawczyk\footnotemark[2] \\
  RIT \\
  Rochester NY, USA \\
  \texttt{bartosz.krawczyk@rit.edu} \\
  \And
  Michał Woźniak\footnotemark[1] \\
  Wrocław Tech \\
  Wrocław, Poland \\
  \texttt{michal.wozniak@pwr.edu.pl} \\
}

\begin{document}

\maketitle

\begin{abstract}

 Memorization impacts the performance of deep learning algorithms. Prior works have studied memorization primarily in the context of generalization and privacy. This work studies the memorization effect on incremental learning scenarios. Forgetting prevention and memorization seem similar. However, one should discuss their differences. We designed extensive experiments to evaluate the impact of memorization on continual learning. 
We clarified that learning examples with high memorization scores are forgotten faster than regular samples. Our findings also indicated that memorization is necessary to achieve the highest performance. However, at low memory regimes, forgetting regular samples is more important. We showed that the importance of a high-memorization score sample rises with an increase in the buffer size. 
We introduced a memorization proxy and employed it in the buffer policy problem to showcase how memorization could be used during incremental training. We demonstrated that including samples with a higher proxy memorization score is beneficial when the buffer size is large.


\end{abstract}

\section{Introduction}

Memorization plays an important role in deep learning 
\cite{feldman2021doeslearningrequirememorization,feldman2020neuralnetworksmemorizewhy}. Deep learning models are capable of memorizing the entire dataset with completely random training labels \cite{zhang2017understandingdeeplearningrequires}, achieving almost perfect training accuracy. 
Methods that limit memorization, such as privacy-preserving algorithms \cite{papernot2017semisupervised}, fail to achieve classification accuracy comparable to that of standard training methods.
Motivated by these observations, later works have shown that achieving high test set performance requires memorization \cite{feldman2021doeslearningrequirememorization}. 
So far, memorization has been studied primarily through the lens of generalization or privacy \cite{wei2024memorizationdeeplearningsurvey}.
Continual Learning (CL) \cite{Chen:2018} is the area of machine learning that aims to prevent catastrophic forgetting \cite{catastrophic_forgetting} in the models trained incrementally with data sampled from different data distributions. 
There were many methods developed that aim to prevent forgetting \cite{masana2022classincrementallearningsurveyperformance}. 
Counterintuitively, memorization and forgetting prevention are not the same. According to the definition of Feldman \cite{feldman2020neuralnetworksmemorizewhy}, memorization refers to the phenomenon where some sample is classified correctly when it is included in the training data. Once removed from the training dataset, it is no longer properly classified. It means that the model can only memorize the sample label without learning any pattern. Such phenomena could occur when the sample has an incorrect label due to a labeling error or when the sample belongs to a long tail \cite{feldman2021doeslearningrequirememorization} (meaning it could be a minority class in an imbalanced setting or it could contain weakly represented features in the dataset). 
The opposite of forgetting is not memorization, but complete knowledge retention \cite{Chen:2018}. In this case, we are interested in keeping the predictive performance high on past tasks, regardless of whether the sample belongs to the long tail or not. 
While forgetting and memorization are different, studying them requires understanding how knowledge is retained in the network.
We found that there are relatively few works \cite{jagielski2023measuring,maini2022characterizingdatapointssecondsplitforgetting,tirumala2022memorizationoverfittinganalyzingtraining} that study the connection between memorization and forgetting. Moreover, these studies examine forgetting in the context of privacy or label noise, with no focus on the change in data distribution.
For this reason, we want to study what is the role of memorization in an incremental training setup, and how Continual Learning is impacted by memorization of training data.

Our key findings could be summarized as follows:
\begin{itemize}
    \item Increasing the number of classes in dataset increases memorization. 
    \item The higher the memorization score of an example, the greater its susceptibility to forgetting.
    \item When training with full access to data from past tasks, the classification accuracy of memorized samples remains high.
    \item A computationally efficient proxy for memorization score can effectively guide buffer policies to improve the performance of incremental training.
    \item The importance of examples, with a significant proxy memory score, increases for large buffers. 
\end{itemize}


\section{Related Works}

\subsection{Memorization}

Memorization \cite{zhang2017understandingdeeplearningrequires} in deep learning refers to a phenomenon where deep neural networks learn specific details or particular features of individual training examples, rather than extracting common patterns or generalized features of the underlying data distribution. This can lead to a model's reliance on rote recall, potentially impacting its ability to generalize to new, unseen data and raising concerns about security and privacy \cite{wei2024memorizationdeeplearningsurvey}.
In \cite{zhang2017understandingdeeplearningrequires}, it was shown that overparametrized neural networks are capable of fitting training data with random labels with high training accuracy. Further work \cite{jastrzebski} studied the impact of architecture and dataset size on memorization. It also shows that general concepts are learned early in training, and memorization could lead to more complicated decision boundaries. 
Anagnostidis et al. \cite{anagnostidis2023the} showed that even with random labels, deep neural networks learn some features that are beneficial for classification on data with original labels. 
Further works \cite{10.5555/3361338.3361358,10.5555/3618408.3619391,tirumala2022memorization} show that memorization does not necessarily lead to overfitting.

In \cite{feldman2021doeslearningrequirememorization}, it was shown that obtaining high accuracy on the test set requires memorization. The atypical samples from the long tail are too few to enable proper representation learning, leading to memorization as the only way of obtaining high performance on the test set.
In \cite{feldman2020neuralnetworksmemorizewhy} Feldman et al. introduced an influence score that allows for estimating the influence of training samples on test samples predictions, finding that there are training examples that impact the correct classification of unseen long-tail data. 

So far, there is no consensus among researchers on what parts of the neural network are responsible for memorization \cite{wei2024memorizationdeeplearningsurvey}. Some evidence \cite{anagnostidis2023the} suggested that the last layers are used for memorization. Other studies \cite{10.5555/3618408.3619391} 
indicated that neurons responsible for memorization are scattered across the whole network. In \cite{feldman2020neuralnetworksmemorizewhy}, it was shown that a classifier is not responsible for memorization, suggesting that memorization is a phenomenon that concerns primarily representation learning.

\subsection{Continual Learning}

Continual learning addresses the challenge of training models on a sequence of tasks with differing data distributions, rather than on a single i.i.d. dataset \cite{Chen:2018}. A major issue in this setting is catastrophic forgetting, where performance on earlier tasks sharply deteriorates as the model learns new ones \cite{catastrophic_forgetting}. 
To tackle this, continual learning methods are generally grouped into three categories. Regularization-based methods reduce forgetting by constraining changes to important parameters. Elastic Weight Consolidation (EWC) adds a regularization term to penalize updates to critical weights \cite{DBLP:journals/corr/KirkpatrickPRVD16}. Learning without Forgetting (LwF) maintains previous knowledge by using pseudo-labels from earlier task classifiers \cite{DBLP:journals/corr/LiH16e}. 
FeTrIL \cite{petit2023fetrilfeaturetranslationexemplarfree} applies a pseudo-feature generation strategy with the usage of a frozen backbone.
Magistri et al. \cite{DBLP:conf/iclr/MagistriTS0B24} address feature drift by regularizing direction relevant for past tasks.

Rehearsal-based methods rely on memory buffers to replay examples from previous tasks \cite{chaudhry2019tinyepisodicmemoriescontinual}. Gradient Episodic Memory (GEM) and its efficient variant aGEM, constrain gradient updates to prevent loss on earlier tasks \cite{DBLP:journals/corr/abs-1812-00420,DBLP:journals/corr/Lopez-PazR17}. More recent work addresses the limitations of small memory buffers using asymmetric updates and classifier corrections \cite{chrysakis2023online}. Other strategies, like Dark Experience Replay (DER), store model logits alongside data and use them in a distillation loss to preserve knowledge \cite{DBLP:journals/corr/abs-2201-00766,buzzega2020dark}. 
Buffer policy algorithms select what samples should be stored in the buffer \cite{hao2023bilevel,9880055,tong2025coreset}. Researchers used bilevel optimization \cite{hao2023bilevel,tong2025coreset}, gradient approximation \cite{10.5555/3454287.3455345,9880055}, or classification uncertainty \cite{9577808} to select samples stored in the buffer.

Expansion-based methods adapt the model architecture to accommodate new tasks. Progressive Neural Networks (PNNs) add task-specific subnetworks that reuse prior knowledge through lateral connections \cite{DBLP:journals/corr/RusuRDSKKPH16}. Other approaches expand network parameters with selective retraining to retain performance on old tasks \cite{DBLP:journals/corr/abs-1708-01547}. Some methods further enhance this by introducing additional convolutional features and training with a specialized loss function to encourage diverse representations for new data \cite{DBLP:journals/corr/abs-2103-16788}.

\subsection{Memorization in Continual Learning}

Memorization in Continual Learning was studied primarily through the lens of privacy 
\cite{ozdenizci2025privacyaware,tobaben2025combinedifferentialprivacycontinual}. In \cite{desai2021continuallearningdifferentialprivacy}, Differential Privacy 
for Continual Learning was proposed. This algorithm uses a data sampling strategy and moment accountant to provide formal privacy guarantees across tasks, achieving tighter privacy loss while maintaining model utility. Authors of \cite{tobaben2025combinedifferentialprivacycontinual} use a prototype classifier with adapters to ensure that the pretrained model can be continually trained with improved privacy. 
Prior works have evaluated the connection between forgetting and memorization \cite{jagielski2023measuring,maini2022characterizingdatapointssecondsplitforgetting,tirumala2022memorizationoverfittinganalyzingtraining}. However, this research was conducted on data with stationary distribution. 

\section{Methods}

\subsection{Notation and Setting}

In continual learning, a neural network is trained on a sequence of tasks with different data distributions. Each task $t$ is defined by a dataset $D_t = \{ (x_i, y_i) \}_{i=0}^{n_t}$, where $x_i$ is image, $y_i$ is label, and $n_t=|D_t|$. We focus on the Class-Incremental Learning setting \cite{vandeven2019scenarioscontinuallearning}, where each task consists of a disjoint set of classes. 
The model $f$ with parameters $\theta$ is trained sequentially by minimizing the loss on the current task: $\mathcal{L}(f(\theta), D_t)$, with access only to the most recent data. Rehearsal-based methods maintain a buffer $\mathcal{M} = \{ (x_j, y_j) \}_{j=0}^{m}$, where $m \ll n_t$
 , to store selected examples from previous tasks.
To avoid ambiguity, we refer to training on the full dataset as \emph{stationary training} (with a fixed data distribution), and to sequential task training as \emph{incremental training}.


\subsubsection{Memorization score}

There are many possible definitions of memorization proposed in the literature \cite{wei2024memorizationdeeplearningsurvey}. 
This work adopts the Feldman \cite{feldman2021doeslearningrequirememorization} definition based on memorization score:

\begin{equation}
    mem(i, A) = E_{f \sim A(D)}[P(f(x_i) = y)] - E_{f \sim A(D / x_i)}[P(f(x_i)=y)]
\end{equation}

where $A$ is a training procedure containing randomness that produces a trained network $f$. Memorization score is the difference between the probability assigned to the correct label by the model trained on the whole dataset and the probability assigned by a model with the $i$-th sample removed from the dataset. Such a definition allows for the detection of memorized samples, but it requires training multiple networks for each learning example to compensate for randomness in $A$. For this reason, Feldman et al. \cite{feldman2020neuralnetworksmemorizewhy} introduced an estimator that reduces the required computational burden, defined as:

\begin{equation}
    mem_{k}(i, A) = E_{f \sim A(\mathcal{S}_{i} )}[P(f(x_i) = y)] - E_{f \sim A( \mathcal{S}_{-i} )}[P(f(x_i)=y)] 
\end{equation}

where $\mathcal{S}_{i}$, $\mathcal{S}_{-i}$ are two random subsets of $D$, each of size $k$, with $S_i$ containing $x_i$ and $S_{-i}$ excluding $x_i$.
Feldman showed that this estimator's variance is bounded by a function of $k/n$ and the number of networks trained, $u$. To keep variance small in our experiments we choose $k/n = 0.5$ and $u = 250$. According to Feldman bound \cite{feldman2020neuralnetworksmemorizewhy}, such a setup ensures that $mem_m$ variance is below 0.016. We study the accuracy of this estimator in more depth in Appendix \ref{app:memscore_proxies}. 
Unless stated otherwise, we use reduced ResNet18 model \cite{DBLP:journals/corr/HeZRS15}, tailored to CIFAR100 \cite{Krizhevsky09learningmultiple,DBLP:journals/corr/Lopez-PazR17} dataset. Overall, for the next part of the experiments, we trained over 3500 neural networks.
The exact hyperparameters used for training are given in the Appendix \ref{app:hyperparameters}. When we average the results, we repeat runs with different random seeds, and report standard deviation in tables or as error bars in plots. From now on when referring to memorization score we will mean Feldman estimator.

\subsection{Impact of tasks split on memorization}

\begin{figure}
    \centering
    \includegraphics[width=1.0\linewidth]{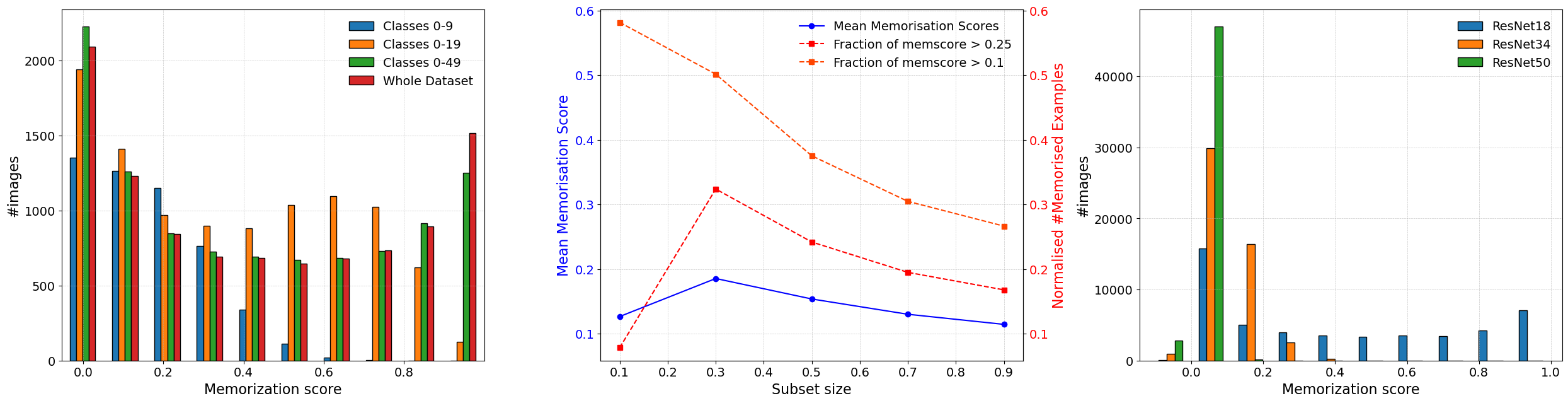}
    \caption{The impact of data and architecture on memorization scores. (Left) histogram of memorization scores for different number classes in the training dataset. (Middle) the depepence between memorization score and number of samples in the dataset. (Right) histogram of memorization scores for different architectures.}
    \label{fig:mem-scores}
\end{figure}

We begin by analyzing how splitting the data into several tasks and changing the number of classes affects the proportion of training samples with high memorization scores.
For this purpose, 
we train ResNet18 \cite{DBLP:journals/corr/HeZRS15} on CIFAR100 \cite{Krizhevsky09learningmultiple} subsets with the first 10, 20, 50, and 100 classes and compute memorization scores for the samples from the first 10 classes. The histogram of memorization scores is presented on the left side of Fig.~\ref{fig:mem-scores}. As the number of classes gets smaller, the memorization scores are lower. This is most evident for score close to one, when full dataset gets largest number of samples.

\begin{wrapfigure}{r}{0.4\textwidth}
    \centering
    \includegraphics[width=0.78\linewidth]{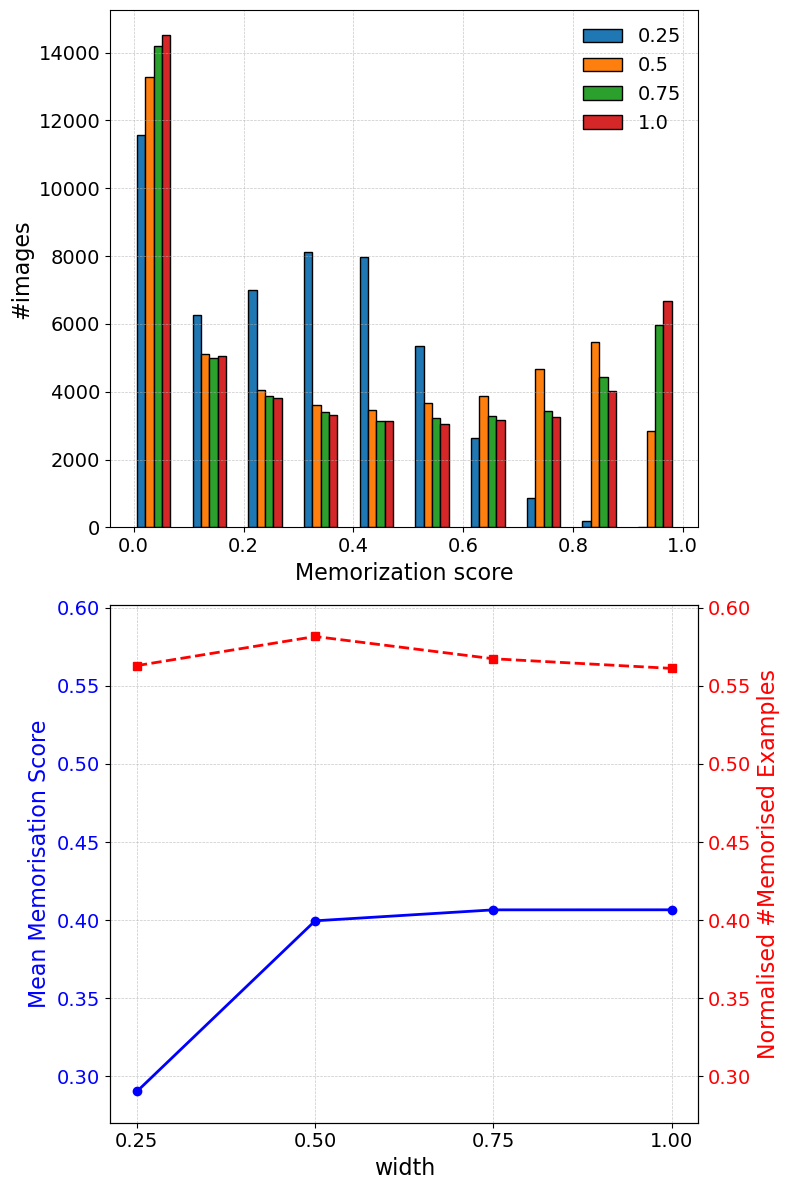}
    \caption{Memorization scores for different model widths}
    \label{fig:memscores-width}
\end{wrapfigure}

Previous results from the literature suggest that reducing dataset size should increase memorization \cite{jastrzebski,10.1109/TIFS.2024.3381477}. By reducing the number of classes, we also limit the number of available training examples. To check if our results are not an artifact of the particular choice of hyperparameters or training regime, we check the impact of dataset size on memorization. To this aim, we evaluate memorization scores for different subsets of CIFAR10 \cite{Krizhevsky09learningmultiple} with 0.1, 0.3, 0.5, 0.7, and 0.9 of the original dataset size. In the middle panel of Fig.~\ref{fig:mem-scores} we plot mean memorization scores and fraction of the dataset with memorization scores above 0.25 and 0.1. Our results are consistent with prior works, suggesting that the number of classes in the dataset has a stronger influence than a reduced number of training samples. One can explain such a phenomenon by the limited capacity of the network. Models trained with a smaller set of classes can learn features that are tailored well to training data, alleviating the need for excessive memorization. If we introduce additional classes into the dataset, while keeping the capacity fixed, the model will have to learn more general features that generalize well across many classes. This interpretation is in line with results from \cite{harun2024what}, where it was shown that a higher number of classes positively correlates with Out-Of-Distribution performance. This means that the model will have to memorize more samples, as the general patterns will not cover more specific samples from all classes. 
To verify our conclusions, we provide in Appendix~\ref{app:additional_memscores} the memorization scores evaluated for 20 and 50 classes, showing very similar trends as for the first 10 classes. We also check if the same trend could be observed in other datasets. We repeat our experiments for the TinyImageNet dataset \cite{Wu2017TinyIC} and find a very similar trend (please refer to Appendix \ref{app:tinyimagenet_memscores} for results).

\paragraph{Impact of architecture.}
The impact of architecture on memorization and Continual Learning has already been studied in the literature. In \cite{jastrzebski}, it was shown that wider networks fit noisy data better. In the context of Continual Learning, studies have found that wider models tend to forget less, while deeper models exhibit greater forgetting \cite{10.5555/3692070.3692733,pmlr-v162-mirzadeh22a}. Motivated by these findings, we also investigate how model depth and width affect memorization scores.
First, we compute memorization scores for ResNet34 and ResNet50 trained on the entire Cifar100 dataset and plot memorization histograms on the right-hand side of Fig.~\ref{fig:mem-scores}. As the network depth increases, the memorization scores decrease significantly, however we need to point out that both ResNet34 and ResNet50 suffered from overfitting in our experiments. In Fig.~\ref{fig:memscores-width}, we plot memorization scores for ResNet18 trained on full Cifar100 with different width multipliers. With the increase in model width, the mean memorization score quickly saturates, but the increase is visible in the samples with the highest memorization score. Therefore, some parallel exists between wider models that are less prone to forgetting and exhibiting more memorization, but the connection is not strong. 

\subsection{Incremental training}

\begin{figure}
    \centering
    \includegraphics[width=1.0\linewidth]{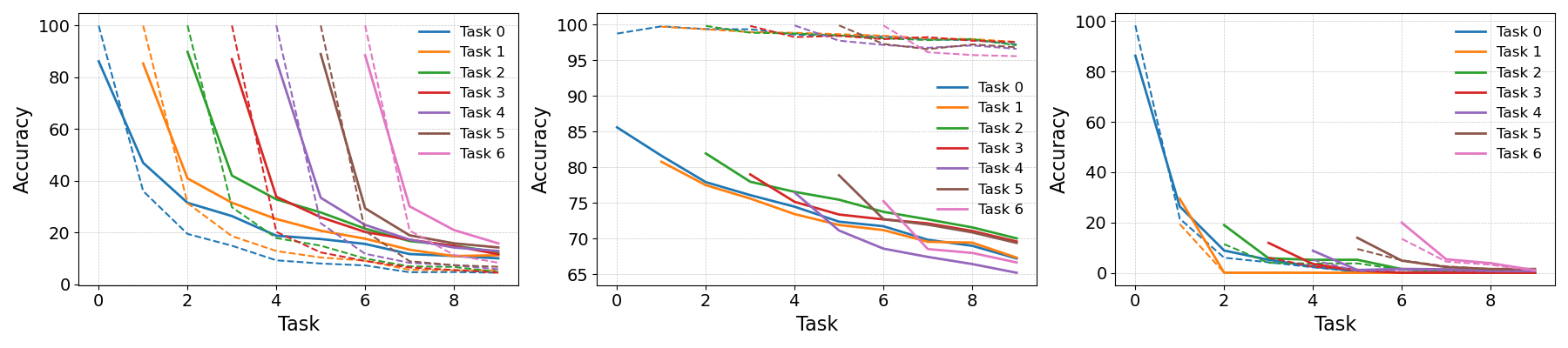}
    \caption{Task accuracy for test set (solid line) and long tail (dotted line) across incremental training on Seq-Cifar100 stream with 10 tasks. (Left) training with buffer size 500. (Middle) training with full access to previous tasks. (Right) training with LwF. Results averaged over 5 runs.}
    \label{fig:memorization-incremental}
\end{figure}

The Feldman formulation of memorization score \cite{feldman2020neuralnetworksmemorizewhy} is difficult to apply in the Continual Learning scenario.
The exclusion of training examples from one task would impact representation learning in the following tasks, leading to a combinatorial explosion of possible training data arrangements that need to be considered when evaluating memorization. To circumvent this problem, we instead use memorization scores computed offline for the whole dataset and track how well samples with higher memorization scores are classified throughout the whole incremental training. We employ a threshold of 0.25 used in \cite{feldman2020neuralnetworksmemorizewhy} to select memorized samples. We study three scenarios. First, we use standard experience replay with reservoir sampling and a buffer of size 500 for Split-Cifar100. As shown on the left-hand side of Fig.~\ref{fig:memorization-incremental}, the accuracy for memorized samples drops significantly after training on two subsequent tasks. Afterward, the accuracy declines at a similar rate for both the test set samples and the training examples with high memorization scores. 
In the second setting, we study training with unlimited access to past tasks' memory (Fig.~\ref{fig:memorization-incremental} middle). During training with an infinite buffer, the accuracy for memorized samples slowly decreases but remains high over the course of the entire training process. This is in line with Feldman's observation about memorization being beneficial and necessary for high performance \cite{feldman2021doeslearningrequirememorization}. It is also in line with our previous results for memorization with a different number of classes in the training set. As we introduce new classes, the model needs to learn more general features, gradually increasing the number of samples that need to be directly memorized. 
Lastly, we study the incremental training with a method that does not use a buffer. We choose Learning without Forgetting \cite{DBLP:journals/corr/LiH16e} (LwF) due to its popularity. On the right-hand side of Fig.~\ref{fig:memorization-incremental}, we see that the accuracy for memorized samples also decreases significantly after the data distribution change. For newly introduced tasks, the accuracy of memorized learning examples is lower than the test set accuracy. In previous scenarios, the initial accuracy for memorized samples was close to 100\%. 
It shows that learning with heavy regularization affects plasticity, reducing the model's capability to classify data with high memorization scores well. 
On the other hand, the accuracy for memorized samples does not drop entirely to zero instantly, which suggests that some trace of memorized data remains in the network, even without access to past data. In Appendix \ref{app:memorization_thresholds}, we include results with different thresholds for determining memorized samples and show that a change in threshold value does not affect our conclusions significantly.


\subsection{Incremental representation learning}
\label{sec:hess_representation_learning}

\begin{wrapfigure}{r}{0.4\textwidth}
    \centering
    \includegraphics[width=1.0\linewidth]{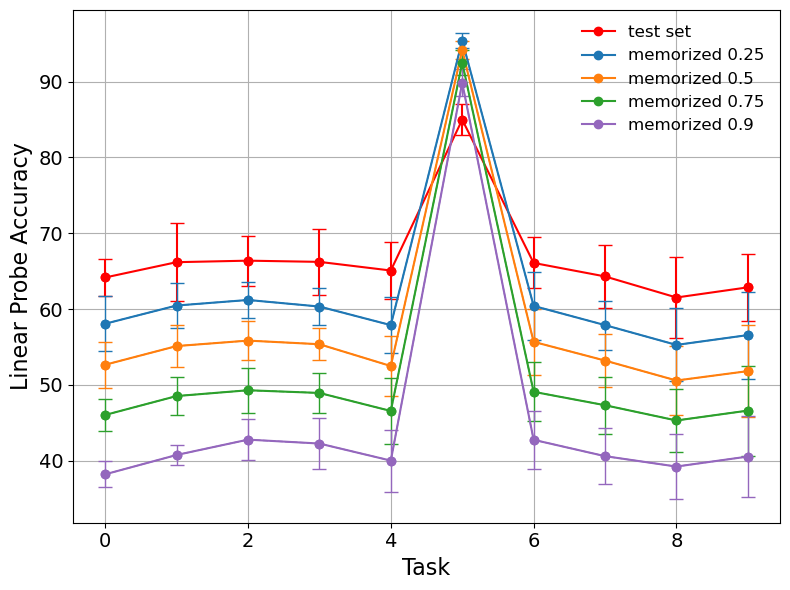}
    \caption{Linear probe accuracy for task 5 during incremental training with SGD on Seq-Cifar100. Results averaged over 5 runs.}
    \label{fig:linear_probes}
\end{wrapfigure}

Hess et al. \cite{hess2024knowledge} analyzed forgetting in representations with the usage of linear probes (LP) to show that forgetting in representation should be considered in reference to performance before and after the training with a given task.
Here we reproduce the experiment from \cite{hess2024knowledge} by training linear probes with data from task 5 on frozen representations obtained from ResNet18 trained incrementally with SGD on Seq-Cifar100. We use hyperparameters reported in the original papers. Fig.~\ref{fig:linear_probes} shows the same trend as the original paper - the LP test accuracy quickly drops to values obtained before training with a given task.

Motivated by our findings with LwF, we study the accuracy of memorized data for the setting introduced by Hess et al. \cite{hess2024knowledge}. Similar to the previous experiment, we select data from the training set with a memorization score exceeding some threshold and evaluate LP with it. Fig.~\ref{fig:linear_probes} shows the same dynamic for training and test data, even with the highest memorization threshold. Previous work on memorization in stationary training \cite{feldman2020neuralnetworksmemorizewhy} showed that memorization is present in representation and not in the classifier. In our experiment, LP trained on representations from other tasks can correctly classify a substantial portion of the data, even with the highest memorization. It suggests that data with high memorization scores computed offline doesn't necessarily correspond to data that is actually memorized in incremental training.
It means we must be careful with our analysis, as determining what is remembered during incremental analysis requires a different method.


\subsection{Memorization score proxy}

\begin{wrapfigure}{r}{0.4\textwidth}
    \centering
    \includegraphics[width=1.0\linewidth]{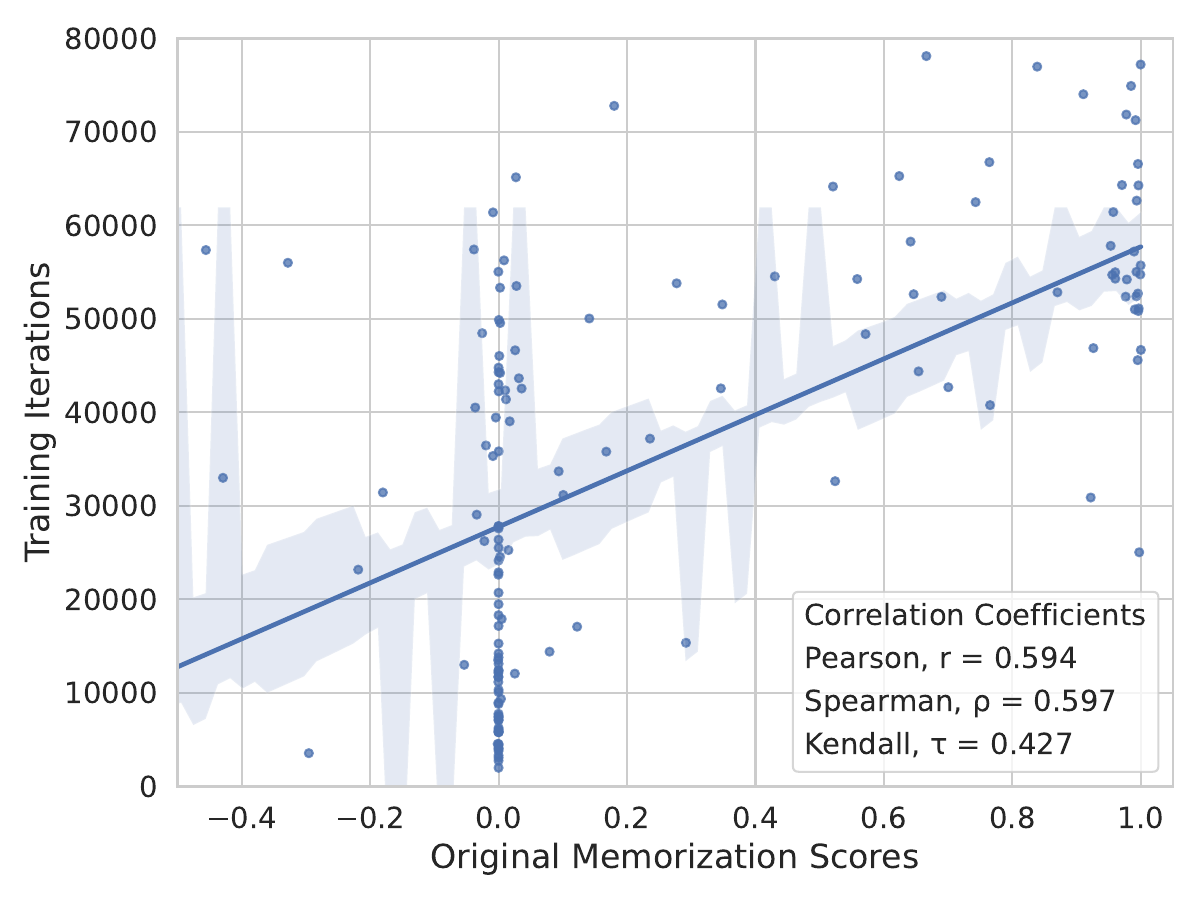}
    \caption{Correlation of training iteration with memorization score.}
    \label{fig:correlation}
\end{wrapfigure}

Determining the memorization score is compute-intensive, even using Feldman estimator \cite{feldman2020neuralnetworksmemorizewhy}. For this reason, it is not feasible to use the memorization score directly during incremental training. To circumvent this issue, we introduce a proxy that approximates the memorization score and is more accessible during training. We base our proxy on the observation made in \cite{jastrzebski}, namely that learning of patterns takes place in early stages of training, while memorization is prone to happen in the later stages of learning. The same premise was used to detect noisy labels in \cite{maini2022characterizingdatapointssecondsplitforgetting}. During offline continual training, we can store the first iteration when the given sample was classified correctly, and later, the prediction did not change in the following epochs. To be more specific, we define our proxy as:

\begin{equation}
    mem_{proxy}(i,A) = min\{ j \in \mathbb{N} | f_j(x_i)=y_i \land \forall_{k > j} f_k(x_i)=y_i \}
\end{equation}

where $f_j$ is the model before the gradient update iteration $j$. In practice, evaluating each sample in every iteration would add a huge overload. For this reason, we check if current predictions match the ground truth after the forward pass with the current minibatch.
This approach is straightforward and adds minimal computational overhead. As shown in Fig.~\ref{fig:correlation}, the correlation between training iteration and original memorization score is moderate, however, we show in Appendix \ref{app:memscore_proxies} that it correlates well with the Feldman estimator.

\subsection{Memorization-aware Experience Replay}

We want to examine how memorization could be used during incremental training. In particular, we are interested in the buffer policy problem - deciding what samples should be stored in the buffer. Storing samples with different memorization scores could impact the overall performance differently. We compute a proposed memorization score proxy to select samples that should correspond to low, medium, or high memorization and use them to update the buffer. The full algorithm for our approach is given in Appendix~\ref{app:full_algo}. 

We are using reservoir sampling and replacing buffer content with selected top-k, mid-k, or bottom-k samples at the end of training. We found in preliminary experiments that such an approach works better than updating the buffer only at the end of task training. 
During the buffer update, we replace the samples from the current task that were already stored in the buffer. We use a balanced variant of reservoir sampling to ensure that the data in the buffer is always balanced.

\section{Experimental setup}

\noindent \textbf{Datasets.} Our experiments follow the class-incremental learning scenario \cite{vandeven2019scenarioscontinuallearning}, using standard continual learning benchmarks created by splitting datasets into multiple tasks. Specifically, we use CIFAR-10, CIFAR-100 \cite{Krizhevsky09learningmultiple}, and Tiny ImageNet \cite{Wu2017TinyIC}, divided into 5, 10, and 20 tasks, respectively. The order of classes across tasks is randomized using different seeds. 

\smallskip
\noindent \textbf{Metrics.} We use two evaluation metrics, namely the final test set accuracy averaged over all tasks, defined as $Acc = \frac{1}{K} \sum_t^K \frac{1}{n_t} \sum_{i=1}^{n_t} \mathbbm{1}[f(x_i, \theta_K) = y_i]$, where $K$ is the number of tasks, and $\mathbbm{1}$ is an indicator function and forgetting measure (FM) \cite{DBLP:journals/corr/abs-1801-10112} defined as average difference between maximum obtained accuracy, and final accuracy for given task.

\smallskip
\noindent \textbf{Baselines.} We use standard reservoir sampling \cite{chaudhry2019tinyepisodicmemoriescontinual}, ballanced reservoir sampling from \cite{DBLP:journals/corr/abs-2010-05595}, Rainbow Memory \cite{9577808}, Bilevel Coreset Selection (BCSR) \cite{hao2023bilevel}, and Probabilistic Bilevel Coreset Selection (PBCS) \cite{pmlr-v162-zhou22h} as baselines in our experiments. For BCSR and PBCS, we use a small auxiliary convolutional model for sample selection, which consists of two convolution layers, a max pool between convolutions, and two linear layers.

\smallskip
\noindent \textbf{Implementation.} 
For all rehearsal-based methods, we use a buffer of size 500 unless specified otherwise. 
When available, we adopt the best hyperparameters reported by the original authors; otherwise, we use the settings detailed in Appendix \ref{app:hyperparameters}. All experiments are implemented using the Mammoth library \cite{buzzega2020dark}.
We made our code available online\footnote{\url{https://github.com/jedrzejkozal/memorization-cl}}.

\section{Results}

\subsection{Evaluation with standard benchmarks}

\begin{table}
    \centering
    \footnotesize
    \caption{Accuracy for incremental training on Split-Cifar10, Split-Cifar100 and Split-TinyImageNet benchmarks. The results averaged over 5 runs.}
    \label{tab:cifar-results}
    \begin{tabular}{ccccccc}
        \toprule
        \multirow{ 2}{*}{buffer policy} & \multicolumn{2}{c}{Split-Cifar10} & \multicolumn{2}{c}{Split-Cifar100} & \multicolumn{2}{c}{Split-TinyImageNet} \\
        \cmidrule(r){2-7}
         & Acc($\uparrow$) & FM($\downarrow$) & Acc($\uparrow$) & FM($\downarrow$) & Acc($\uparrow$) & FM($\downarrow$) \\
        \midrule
         reservoir \cite{chaudhry2019tinyepisodicmemoriescontinual}  & 55.18±6.59 & 42.45±7.04 & 22.07±0.84 & 65.47±0.62 & 6.38±0.62 & 75.87±0.54 \\
         reservoir balanced \cite{DBLP:journals/corr/abs-2010-05595} & 55.25±4.66 & 42.35±4.94 & 22.39±0.75 & 65.01±0.20  & 6.53±0.58 & 75.99±0.68 \\
         rainbow memory \cite{9577808}     & 56.86±3.17 & 39.91±4.07 & 23.19±0.79 & 64.16±0.48 & 6.42±0.52 & 76.07±0.73 \\
         BCSR \cite{hao2023bilevel}               & 55.83±4.71 & 41.37±5.69 & 22.84±0.84 & 64.80±0.53  & 6.65±0.52 & 75.86±0.43 \\
         PBCS \cite{pmlr-v162-zhou22h}              & 55.22±5.31 & 42.35±5.44 & 23.14±1.09 & 64.39±0.89 & 6.93±0.57 & 75.79±0.42 \\
        \midrule
         bottom-k memscores & 56.04±4.80  & 41.80±5.25  & 25.88±1.14 & 61.60±0.71  & 7.85±0.29 & 74.87±0.54 \\
         middle-k memsocres & 55.96±4.62 & 41.80±4.98  & 23.62±1.07 & 63.81±0.54 & 7.11±0.57 & 75.52±0.85 \\
        top-k memscores    & 41.16±7.38 & 56.60±7.85  & 17.28±1.19 & 69.95±0.87 & 5.18±0.13 & 77.12±0.63 \\
        \bottomrule
    \end{tabular}
    \label{tab:benchmarks}
\end{table}

First, we carry out an experiment on a standard set of benchmarks used in Continual Learning. We compare accuracy and FM for the proposed approach with other buffer policy methods in Tab.~\ref{tab:benchmarks}. In the standard setup, we can see that selecting lower and mid proxy memorization scores obtains better results than reservoir sampling. This suggests that with the lowest memory budgets, retaining the performance for standard data is challenging enough, therefore we should be constructing a buffer with the most typical samples that are easy to learn and represent well, given the class. This shows that in such a setting, memorization does not play a significant role as forgetting prevention is more important, however the memorization could be used to guide the buffer construction process.

\begin{table}
\parbox{.50\linewidth}{
    \centering
    \tiny
    \caption{The Accuracy and Forgetting Measure for the ER-ACE and DER++. The results averaged over 5 runs.}
    \begin{tabular}{ccccc}
    \toprule
    \multirow{ 2}{*}{method} & \multicolumn{2}{c}{Split-Cifar100} & \multicolumn{2}{c}{Split-TinyImageNet} \\
    \cmidrule(r){2-5}
                     & Acc($\uparrow$) & FM($\downarrow$) & Acc($\uparrow$) & FM($\downarrow$) \\
    \midrule
     ER-ACE \cite{Caccia:2022}          & 36.81±0.86 & 34.02±0.36 & 15.07±0.35 & 39.68±1.17 \\
           +bottom-k & 40.03±0.29 & 32.18±1.2  & 19.27±0.81 & 35.59±0.76 \\
           +mid-k    & 38.69±0.76 & 31.29±1.08 & 15.76±0.52 & 39.43±1.55 \\
           +top-k    & 33.22±0.69 & 37.84±1.54 & 12.92±0.37 & 42.07±1.42 \\
     \midrule
     DER++ \cite{buzzega2020dark}           & 33.9±1.67  & 50.12±2.16 & 12.5±0.94  & 58.29±1.99 \\
          +bottom-k  & 38.34±1.40  & 45.13±1.42 & 15.09±0.81 & 55.14±1.88 \\
          +mid-k     & 34.75±2.37 & 49.04±2.51 & 13.04±1.26 & 57.80±2.27  \\
          +top-k     & 28.32±2.92 & 56.9±3.07  & 7.60±1.10    & 66.79±2.57 \\    
    \bottomrule
    \end{tabular}
    \label{tab:rehersal_advanced}
}
\hfill
\parbox{.45\linewidth}{
    \centering
    \tiny
    \caption{Test accuracy for buffer policies based on mixing high memorization samples with mid and bottom on Seq-Cifar100 and various buffer sizes.}
    \begin{tabular}{ccc}
\toprule
\multirow{ 2}{*}{method} & \multicolumn{2}{c}{buffer size} \\
\cmidrule(r){2-3}
  & 2000 & 5000 \\
\midrule
 bottom-k memscores & 42.24±0.45                           & 52.64±0.61   \\                 
 +10\% top-k      & 42.17±0.3(\textcolor{Red}{-0.07})    & 52.46±0.62(\textcolor{Red}{-0.18}) \\
 middle-k memsocres & 40.52±0.57                           & 52.75±0.71 \\
 +10\% top-k      & 40.77±0.64(\textcolor{Green}{+0.25}) & 52.74±0.43(\textcolor{Red}{-0.01}) \\
 \midrule
 \multirow{ 2}{*}{method} & \multicolumn{2}{c}{buffer size} \\
 \cmidrule(r){2-3}
 & 10000     & 20000 \\
 \midrule
 bottom-k memscores & 59.56±0.69                           & 64.81±0.87                           \\
 +10\% top-k      & 59.61±0.43(\textcolor{Green}{+0.05}) & 64.91±0.68(\textcolor{Green}{+0.10}) \\
 middle-k memsocres & 59.88±0.52  & 64.57±0.45  \\
 +10\% top-k      & 60.16±0.48(\textcolor{Green}{+0.28}) & 64.93±0.64(\textcolor{Green}{+0.36}) \\
\bottomrule
\end{tabular}
    \label{tab:mixed-strategies}
}
\end{table}

We acknowledge, that our results may lack some of the baselines, that may obtain better performance that the proposed method like \cite{10.5555/3454287.3455345,9880055,tong2025coreset}, however our primary motivations is not to propose the best algorithm for buffer policy, but rather study the impact of memorization on Continual Learning. 
We also show in Tab.~\ref{tab:rehersal_advanced} that the proposed approach also works with other rehearsal algorithms \cite{buzzega2020dark,Caccia:2022}.

\subsection{Training with larger buffers}

\begin{wrapfigure}{r}{0.4\textwidth}
    \centering
    \includegraphics[width=0.95\linewidth]{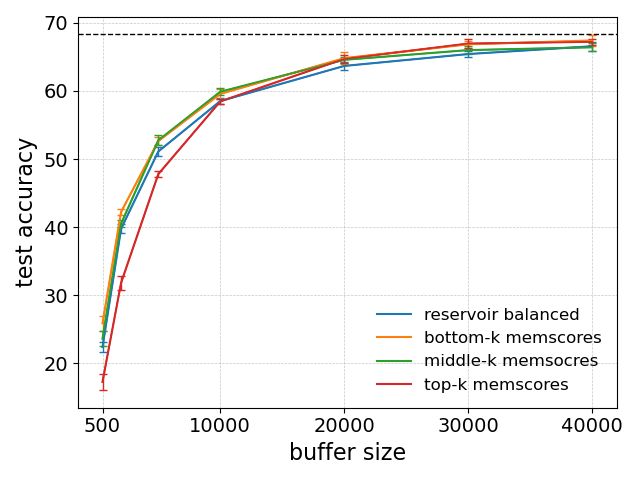}
    \caption{Test set accuracy for various buffer sizes on Split-Cifar100 benchmark. Results are averaged over 5 runs. The black horizontal line denotes training with full access to memory. The results averaged over 5 runs.}
    \label{fig:buffer_sizes}
\end{wrapfigure}

We evaluate how the classification performance of proposed policies changes with an increase in buffer size. In Fig.~\ref{fig:buffer_sizes}, we plot test set accuracies for different buffer policies based on proxy memorization score. Regardless of buffer size, selecting samples with the lowest or medium memorization performs better than random selection. Also, as the buffer size increases, the performance obtained for selecting top-k samples improves. This suggests that keeping hard samples in the buffer is necessary to obtain performance close to the upper bound.

To further evaluate the impact of high memorization in different regimes, we modify the sample selection process for bottom-k and mid-k to reserve 10\% of the current task buffer space for samples with the highest proxy memorization score. The results are presented in Tab.~\ref{tab:mixed-strategies}. As in the previous case, the benefit of including samples with high proxy values is mostly visible for higher buffer sizes. However, the increase in accuracy is lower than the standard deviation, suggesting that the impact is low.

\section{Discussion}

Our results suggest that as the number of classes in the dataset increases, the representations learned by the network become more general. At the same time, larger fractions of long tail 
data not being covered by general patterns should be memorized. We can think of results for each number of classes as the upper bound of performance at different stages of incremental training. It shows the changes in the memorization as we progressively increase the number of tasks.
Increasing memorization could be especially challenging for exemplar-free CL methods, which do not have access to prior labels and keep model capacity fixed. This means that, in principle, Continual Learning with either increasing architecture \cite{DBLP:journals/corr/RusuRDSKKPH16} or rehearsal with memory \cite{buzzega2020dark,chaudhry2019tinyepisodicmemoriescontinual} should be much easier.
We believe that our results also have an impact on the construction of CL benchmarks. The larger number of classes per task means there is a larger need for memorization, while, as we have shown, data with a large memorization score is forgotten faster. On the other hand, a lower number of classes per task requires learning more general representations from a series of more specialized models. In CL benchmarks, the number of classes in the task is usually tight to the number of tasks, as it is limited by the overall number of classes in the used dataset \cite{liu2023benchmarkingcontinuallearningcognitive}. Future benchmarks and evaluations should consider that different numbers of classes are challenging for different reasons. 


The accuracy obtained for selecting samples with the top proxy memorization score increases as the buffer grows. This indicates that when the memory size limitation is relaxed (as it is postulated by several recent Continual Learning works \cite{harun2024grasprehearsalpolicyefficient,10.5555/3524938.3525432,peng2022memory}), the importance of memorized samples is increasing. 
We believe that progress in incremental learning can be achieved by studying both training with high access to memory and low or no memory access. The first facilitates the necessary conditions for achieving high performance, that is, comparable to full access to memory, while the second provides highly efficient methods in terms of memory usage. Practitioners later could select methods from both ends of this spectrum to solve specific problems they work on. 

We acknowledge that our proxy memorization score is similar to prior work on the impact of training order on forgetting \cite{hacohen2024forgettingordercontinuallearning}, however, our results were derived independently with different conclusions. Moreover, that paper does not discuss memorization, which is our main focus.  
Also, the notion of "typical" learning example was used in \cite{harun2024grasprehearsalpolicyefficient} to enhance the replay policy. But we find that the Mahalanobis distance in the representation space is a poor approximation of the memorization score (see Appendix~\ref{app:memscore_proxies}).

\paragraph{Limitations} 
We acknowledge that networks with different architectures could have different memorization dynamics. Memorization could also be affected by the training procedure and hyperparameter selection. Factors that are known to impact memorization include data augmentation \cite{10.1109/TIFS.2024.3381477}, regularization \cite{zhang2017understandingdeeplearningrequires,jastrzebski}, and data repetition \cite{10.5555/3666122.3667830}. In this study, we focused on popular Continual Learning setups, which are commonly used in the literature \cite{buzzega2020dark}. However, we recognize that additional studies are needed to assess the impact of other factors. 
Moreover, Feldman et al. \cite{feldman2020neuralnetworksmemorizewhy} showed that CIFAR100 has some samples that occur both in training and in the test set, leading to increased influence of some atypical samples on the test set. We tried to address that by using different datasets and settings (see Appendices \ref{app:additional_memscores},\ref{app:memorization_thresholds}, and \ref{app:tinyimagenet_memscores}). However, at the same time, CIFAR100 is one of the most popular datasets used for studying Continual Learning. Future experiments should include broader spectra of datasets to obtain more robust results. Finally, as shown in Section~\ref{sec:hess_representation_learning}, our experiments only study the accuracy of samples with memorization score computed with stationary training. We do not solve the problem of determining exactly what samples are memorized during incremental training. We leave it for future research, but we acknowledge that developing such a method could shed more light on the role of memorization in Continual Learning.

\section{Conclusion}

Memorization is a necessary component for achieving a classification performance comparable to training with full memory in Continual Learning. Yet, simultaneously, the capability to correctly classify the samples with high memorization scores drops significantly after a change in data distribution. Our experiments with standard CL benchmarks show that at lower memory regimes, forgetting of regular data is a more important consideration than forgetting of long-tail data. Regardless, we show that the notion of memorization can still be useful in constructing buffer policy, even if data with a high memorization score is not important.
Our further experiments show that closing the gap between stationary and incremental training requires taking memorization into consideration.

Future work could include the localization of which parts of the network are responsible for memorization \cite{anagnostidis2023the,10.5555/3618408.3619391} and designing a proper measure to protect these parts from forgetting in incremental training. We know what factors affect memorization in stationary training. However, the factors that could impact memorization in incremental learning are unknown. We believe that studying these factors could also be beneficial.

\begin{ack}
Jędrzej Kozal and Michał Woźniak were supported by the CEUS-UNISONO programme, which has received funding from the National Science Centre, Poland under grant agreement No. 2020/02/Y/ST6/00037 and the statutory fund of Department of Systems and Computer Networks, Wroclaw Tech.
Jan Wasilewski, Alif Ashrafee, and Bartosz Krawczyk were partially supported by RIT Chester F. Carlson Center for Imaging Science funding no. 67922/15961.


\end{ack}

\bibliography{bibliography}

\newpage

\appendix

\section{Broader Impact}

Our work primarily studies the properties of continual learning algorithms in the context of their performance; therefore, it is hard to predict how presented results will impact particular applications, however we find that two key aspects require attention. Firstly, our work shows that memorization needs to be considered when designing well-performing incremental training algorithms. This claim has a significant impact on data privacy, as it puts responsibility on the user of the incremental-training algorithm, which creates a large buffer to store past data. This user must curate the data used for rehearsal. This process should involve ensuring that the data owner has agreed to particular data usage and that the data is properly licensed and anonymized. Failure to adjust to these requirements can lead to leakage of private or proprietary information used for training.

The second consideration is the higher computational cost compared to other experiments. Evaluating the memorization requires more computational power due to the repeated training process, which comes at increased environmental costs due to carbon emissions. However, the incremental learning has the potential to reduce the required training time by removing the need for training from scratch every time new data with a different distribution arrives. We believe that our experiments can potentially reduce the cost of updating deep learning models in the future, at the expense of larger power consumption now.

\section{Memorization score proxies}
\label{app:memscore_proxies}

The original memorization scores defined in \cite{feldman2021doeslearningrequirememorization} require several neural network trainings to compute a score for only a single sample. For this reason, we analyzed several possible proxies that require less computation and could be applied more easily in our experiments. We have considered the following proxy scores for memorization:

\begin{itemize}
    \item \textbf{Feldman estimator} - introduced in \cite{feldman2020neuralnetworksmemorizewhy}, the estimator limits the number of trainings required to compute the memorization scores. If the original memorization score can be defined as leave-one-out, the estimator is leave-k-out. It requires less computation compared to the original memorization score, however, it is still not usable during incremental training.
    \item \textbf{Mahalanobis distance} - we compute the Mahalanobis distance in the learned latent representation space of the neural network. We use a feature vector representing the given sample and the mean and covariance of the class to which the sample belongs. A similar solution for rehearsal policy was introduced in \cite{harun2024grasprehearsalpolicyefficient}.
    \item \textbf{LASS distance} - in \cite{jastrzebski} the Langevin adversarial sample search (LASS) was introduced to find adversarial samples that are in some predefined $||L||_{\infty}$ neighborhood of the sample. The authors used this method to study how complicated the decision boundary is and found that neural networks trained with random labels have much more complicated decision boundaries compared to standard training. Here we reuse the LASS algorithm, however, we do not limit the neighborhood size. Instead, we set it to some large constant and use distance from the original sample $||x_{adv} - x||_{\infty}$ as the proxy.
    \item \textbf{Carlini-Wagner distance} - the LASS algorithm was designed primarily to better explore the search space and not to find the minimal perturbance that causes a change in classification. For this reason, we use an untargeted Carlini-Wagner adversarial attack \cite{7958570}, that directly optimizes for the smaller perturbance in the $||L||_2$. We utilize this algorithm to generate the adversarial example, and then compute the $||x_{adv} - x||_2$ difference and use it as a proxy.
    \item \textbf{Training iteration} - based on previous results \cite{jastrzebski,maini2022characterizingdatapointssecondsplitforgetting} we can assume that simple patterns are trained in the first epochs, while memorization mostly happens later. For this reason, we may consider the iteration at which the sample was classified correctly (and was classified correctly until the end of the training) as a valid proxy for memorization score.
\end{itemize}

\begin{figure*}[!ht]
    \centering
    \begin{subfigure}[t]{0.32\textwidth}
        \centering
        \includegraphics[width=\textwidth]{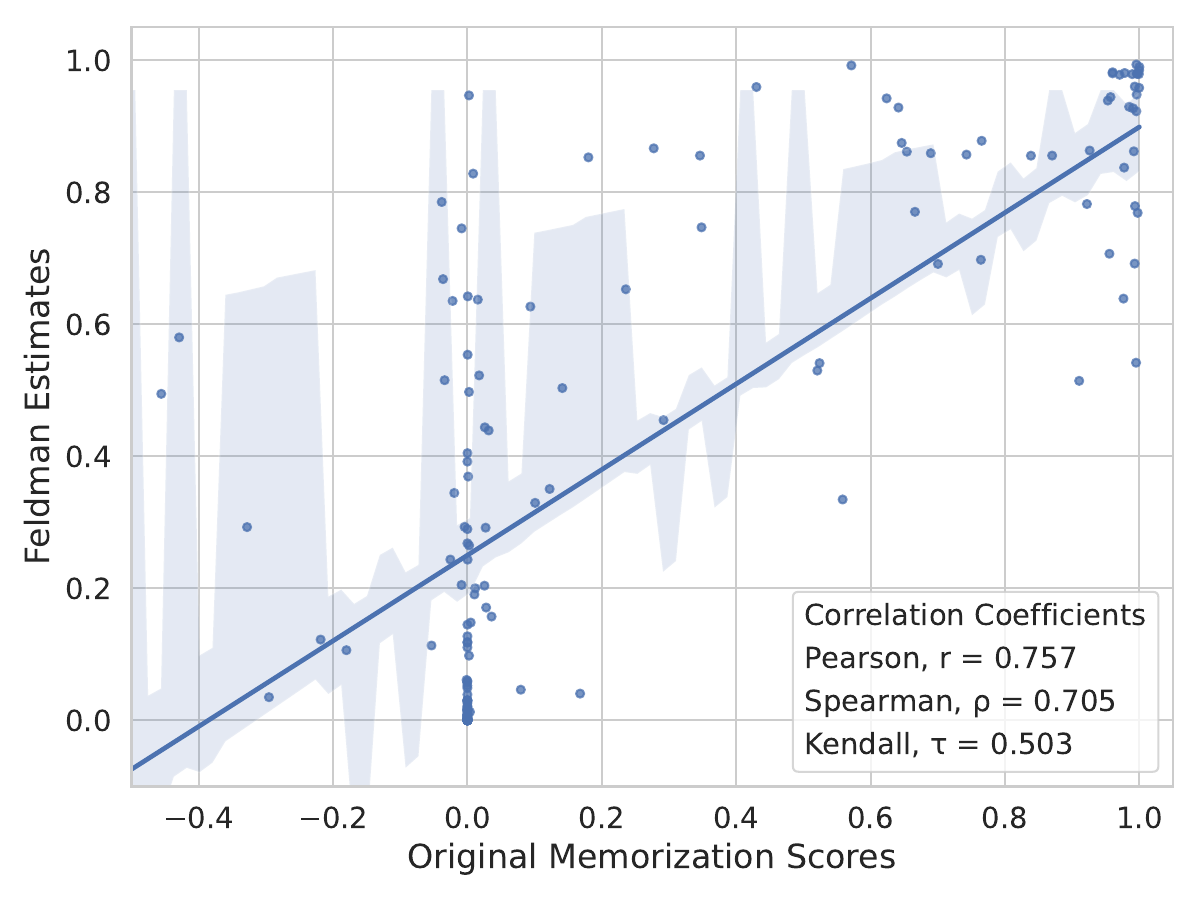}
        \caption{\tiny{Org. Mem. Scores vs Feldman Estimator}}
        \label{fig:original_mem_scores_vs_feldman}
    \end{subfigure}
    \begin{subfigure}[t]{0.32\textwidth}
        \centering
        \includegraphics[width=\textwidth]{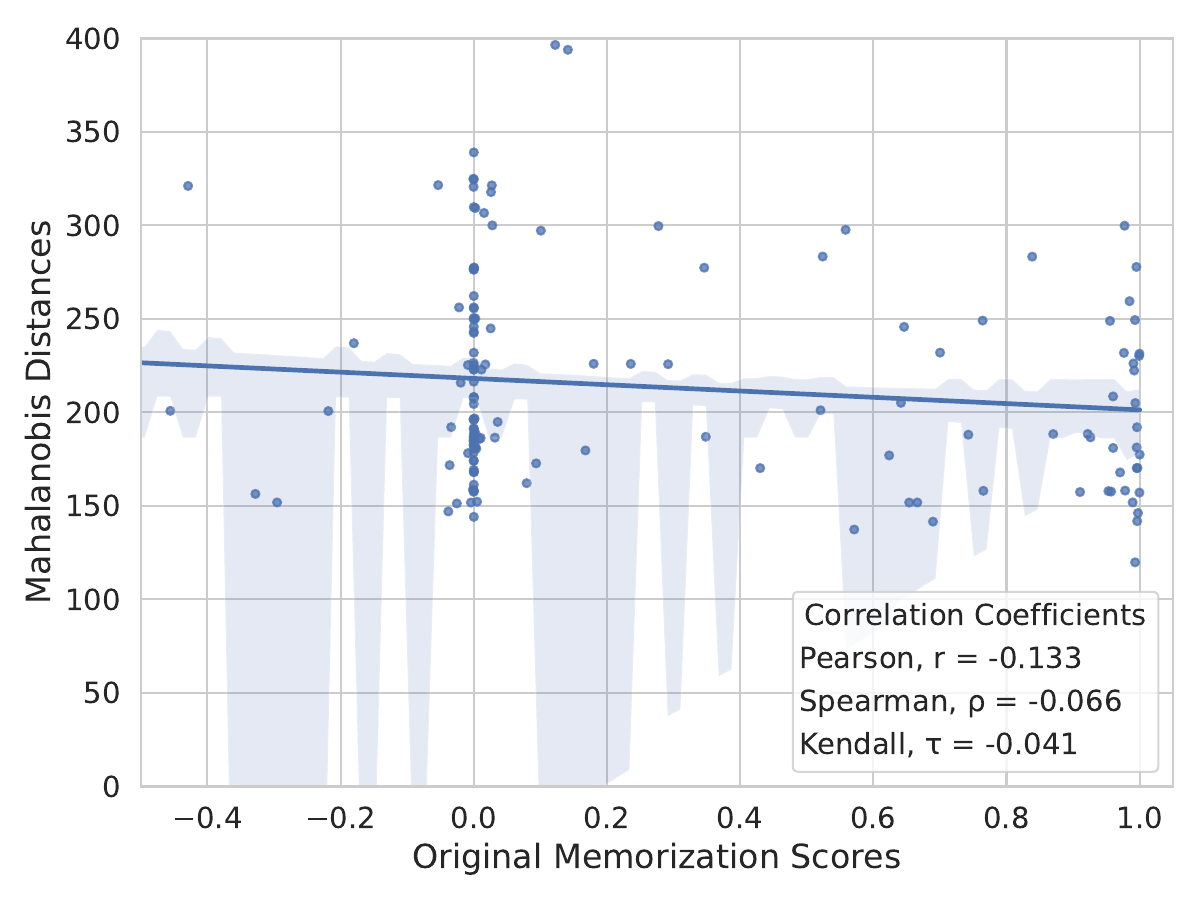}
        \caption{\tiny{Org. Mem. Scores vs Mahalanobis Distance}}
    \end{subfigure}
    \begin{subfigure}[t]{0.32\textwidth}
        \centering
        \includegraphics[width=\textwidth]{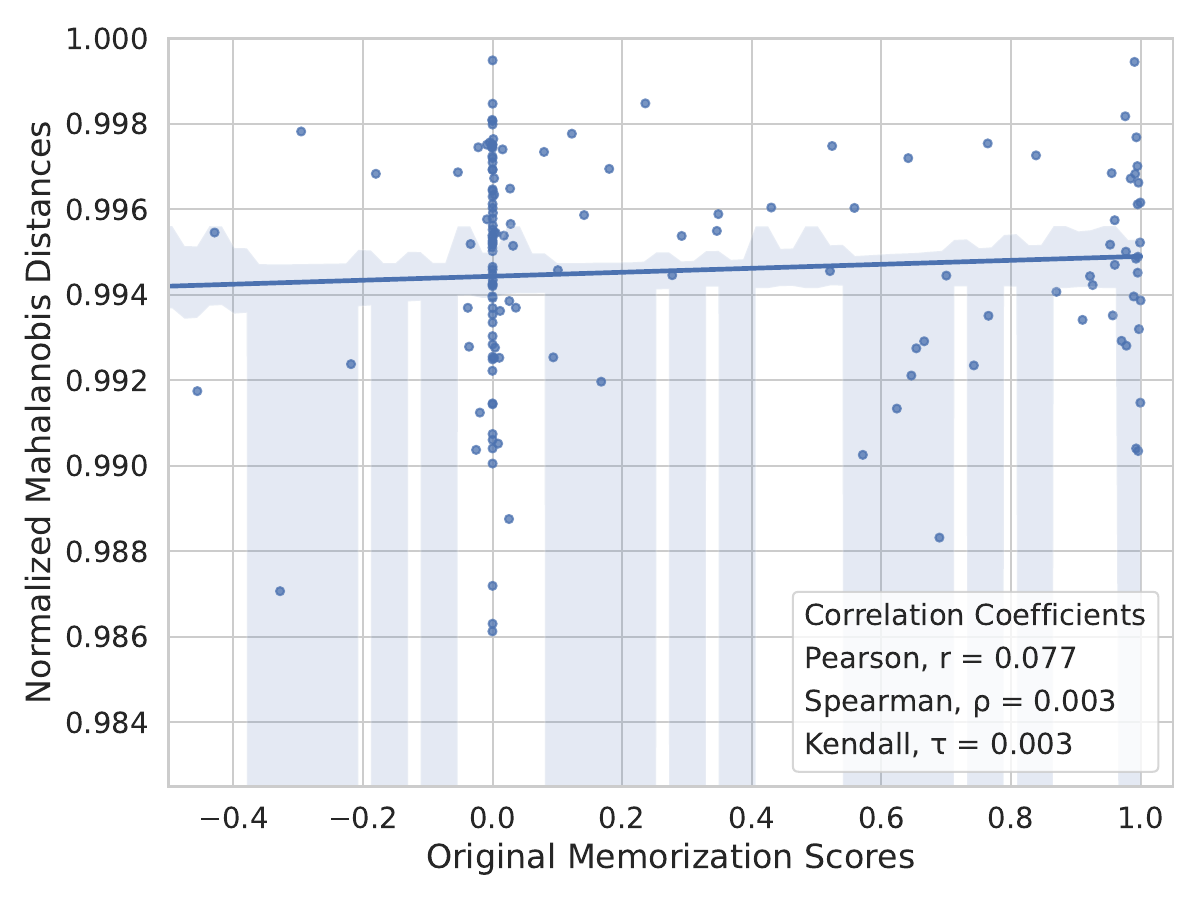}
        \caption{\tiny{Org. Mem. Scores vs Normalized Mahalanobis}}
    \end{subfigure}
    \begin{subfigure}[t]{0.32\textwidth}
        \centering
        \includegraphics[width=\textwidth]{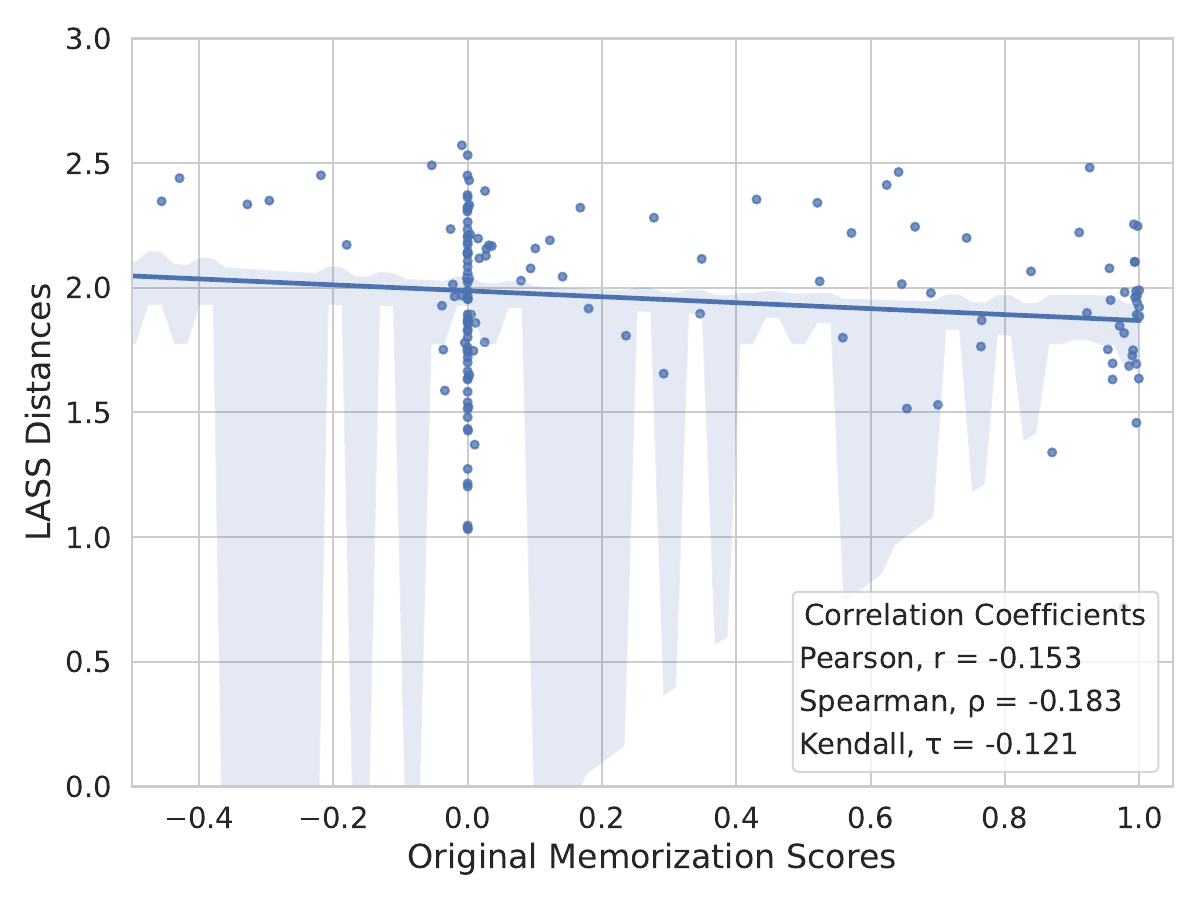}
        \caption{\tiny{Org. Mem. Scores vs LASS Distance}}
    \end{subfigure}
    \begin{subfigure}[t]{0.32\textwidth}
        \centering
        \includegraphics[width=\textwidth]{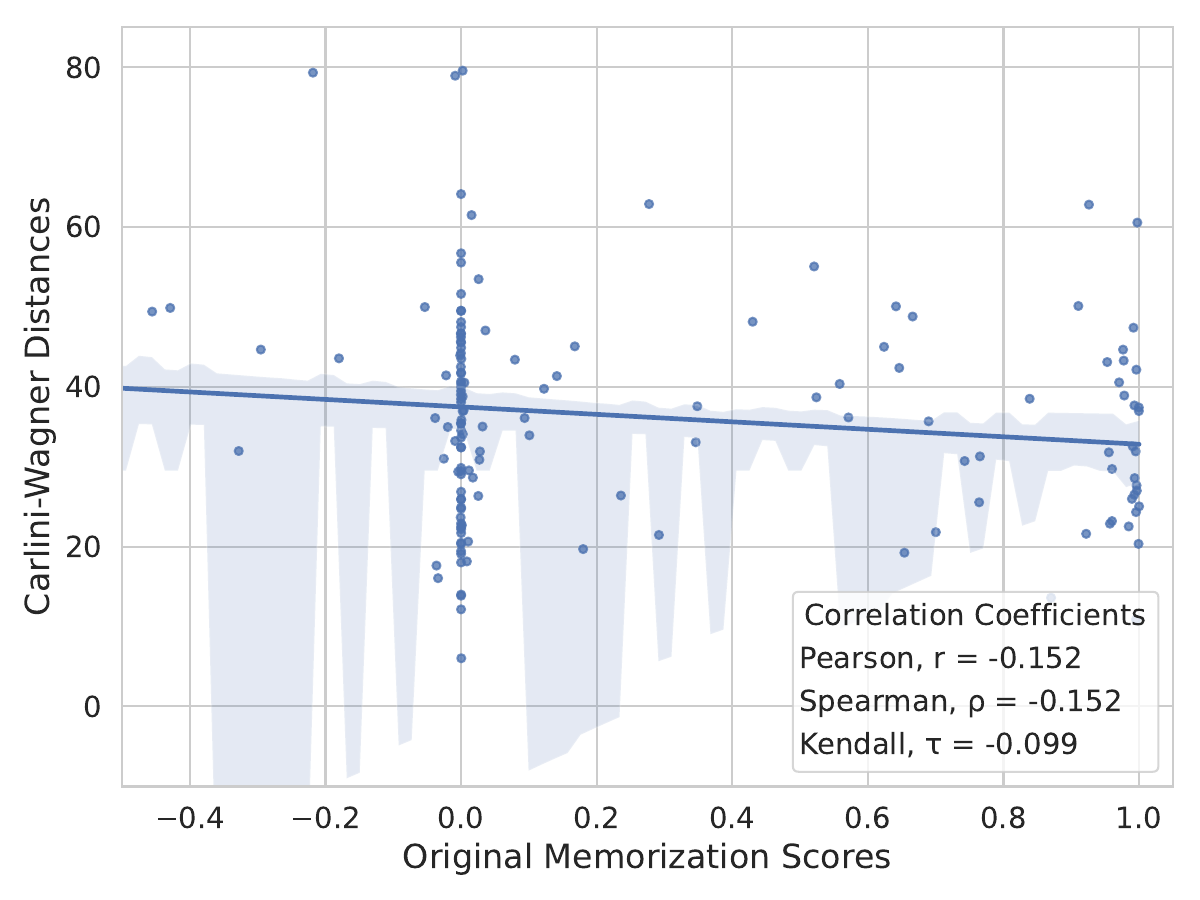}
        \caption{\tiny{Org. Mem. Scores vs Carlini-Wagner Distance}}
    \end{subfigure}
    \caption{Correlation between original memorization scores and various proxies}
    \label{fig:proxy_correlations}
\end{figure*}

To evaluate how closely the proxies match the original memorization scores, we compute the original memorization score for randomly sampled 150 samples from CIFAR100, and then measure the correlations between each proxy and the memorization score (see Fig.~\ref{fig:proxy_correlations}). To limit the computation used during the calculation of the original memorization score, we train only a single network for a single sample excluded from the dataset and a single network for the whole dataset.

For each correlation, we report Pearson $r$, Spearman $\rho$, and Kendall $\tau$. Pearson’s $r$ quantifies the linear relationship between two variables. Spearman’s $\rho$ is a rank‑based measure of monotonic association, reflecting how often the ordering by proxy matches the ordering by the original score. Kendall’s $\tau$ is a more robust ordinal metric, less sensitive to ties and outliers.

\begin{figure}[!ht]
    \centering
        \includegraphics[width=0.48\textwidth]{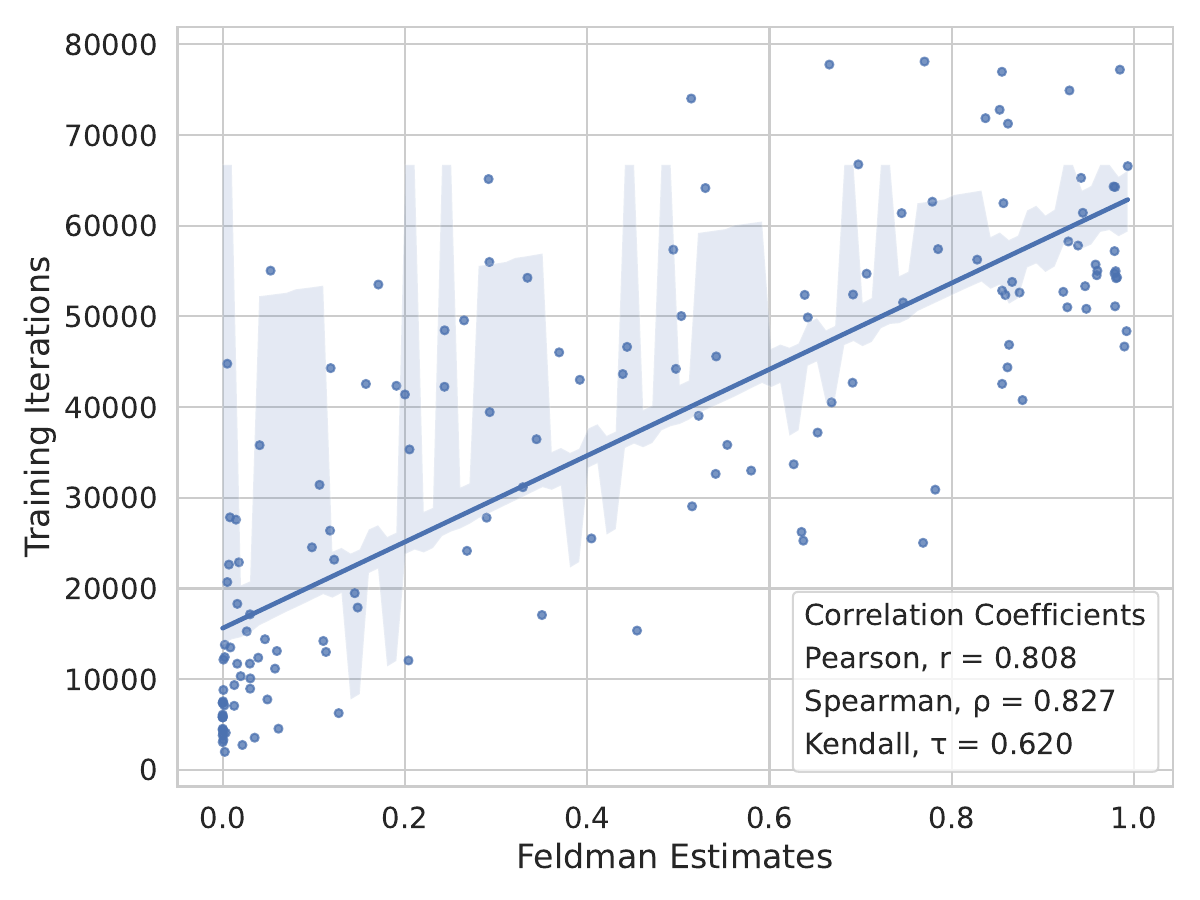}
        \caption{Feldman Estimator vs Training Iteration Estimates}
        \label{fig:feldman_vs_training_iterations}
\end{figure}

Figure~\ref{fig:original_mem_scores_vs_feldman} shows that the original memorization scores and the Feldman estimator are strongly correlated (Pearson $r=0.76$, Spearman $\rho=0.71$, and Kendall $\tau=0.50$), whereas the other proxies exhibit much weaker associations. To demonstrate that our training-iteration proxy follows the same pattern, we compare it directly against the Feldman estimator in Fig.~\ref{fig:feldman_vs_training_iterations}, again observing strong agreement.

Our training-iteration proxy achieves Pearson $r=0.81$, Spearman $\rho=0.83$, and Kendall $\tau=0.62$ against the Feldman estimator, indicating it is a reasonable approximation of the Feldman estimator. Since we already have evidence from Fig. \ref{fig:original_mem_scores_vs_feldman} that the Feldman estimator tightly tracks the original memorization scores, and our proxy tightly tracks the Feldman estimator, we can increase our confidence in the training-iteration proxy. Since we use a single training to compute original memorization scores, our estimate could be biased. Having moderate and high correlation with the original memorization score and the Feldman estimator, we can deduce that samples learned later in training could identify highly memorized learning examples.

\section{Training hyperparameters}
\label{app:hyperparameters}

We use a standard set of hyperparameters that are commonly used for various Continual Learning algorithms \cite{buzzega2020dark, DBLP:journals/corr/abs-2201-00766}. For both the continual and stationary cases, we train with SGD for 50 epochs per task, or in the case of non-incremental training, we train for 50 epochs on the entire dataset. All networks (including baselines) are trained with a learning rate of 0.1, which is divided by 10 at epochs 35 and 45. In case of non-incremental training, we used a weight decay of 1e-06 and momentum equal to 0.9. For continual training, we set both weight decay and momentum to 0.0. For all training setups, we use a batch size of 32 for both the current data and samples from the memory buffer. 
For all experiments with continual learning benchmarks, we provide on our repository a complete registry of our experiment results in MLFlow \cite{10.1145/3399579.3399867} along with hyperparameters used to start each training run. We store each commandline argument used for training as an MLFlow hyperparameter, therefore our results could be easily inspected. We did not keep such a registry for our experiments with the memorization score.

There are two exceptions to the hyperparameters provided above. When training LwF, we used alpha (regularization hyperparameter of knowledge distillation loss) equal to 0.99, learning rate 0.02, 10 epochs, SGD momentum of 0.0, and weight decay 5e-4. We found that this set of hyperparameters works better than the standard set of hyperparameters used for other methods. The second case is the experiments with linear probes, where we used the setting used in the original experiments \cite{hess2024knowledge}, namely: AdamW \cite{loshchilov2018decoupled} optimizer, learning rate of 0.001, weight decay of 0.0005, and batch size of 128.

For all datasets in experiments, we use an augmentation pipeline consisting of the following transformations:
\begin{enumerate}
    \item Random Horizontal Flip with probability 0.5
    \item Random Crop with size of the original image (32 for CIFAR datasets, and 64 for TinyImageNet) and padding of 4
    \item Transformation to tensor - that normalizes image pixels into [0,1] interval
    \item Normalization with channel-wise mean and standard deviation computed for each dataset separately.
\end{enumerate}

This is a standard set of augmentations that is available in Torchvision library.



\section{Additional memorization score results}
\label{app:additional_memscores}

\begin{figure}
    \centering
    \includegraphics[width=1.0\linewidth]{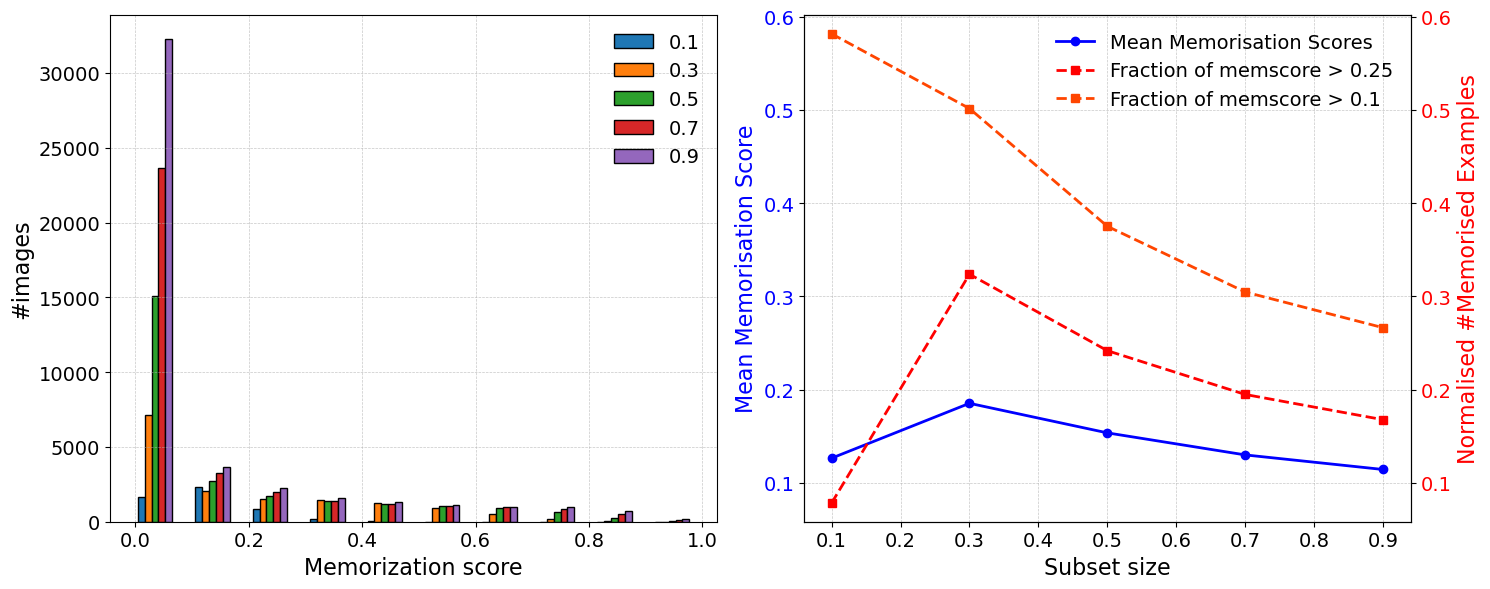}
    \caption{(Left) histogram of proxy memorization scores for different CIFAR10 subsets. (Right) mean memorization score and number of memorized samples normalized by subset size for the data from left plot.}
    \label{fig:apendix_memscores}
\end{figure}

\begin{figure}
    \centering
    \includegraphics[width=0.8\linewidth]{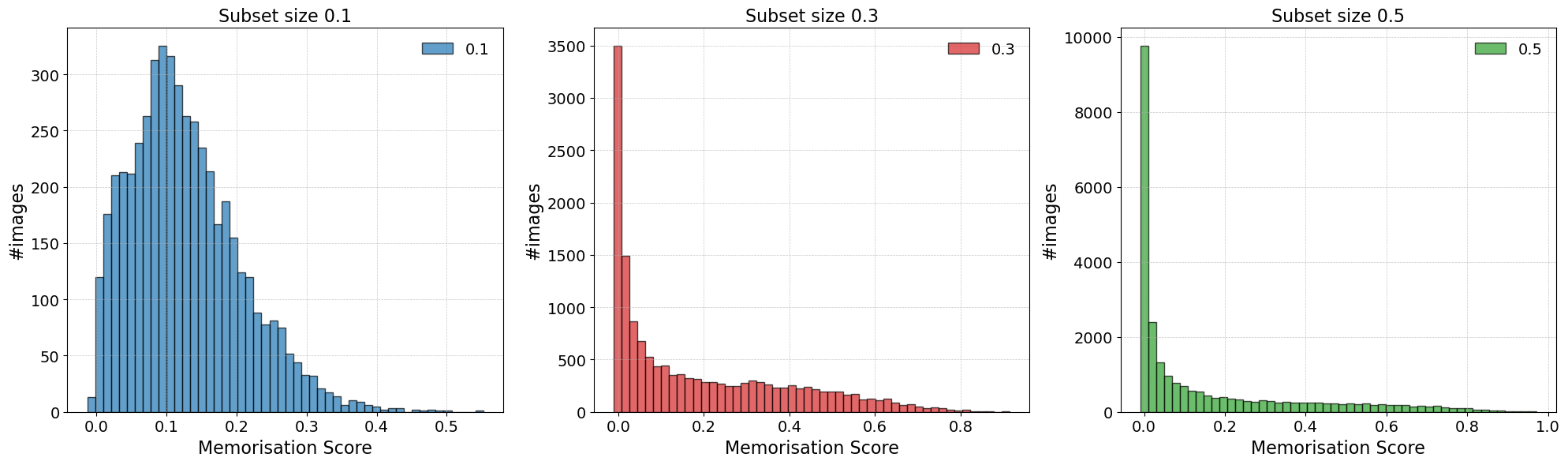}
    \caption{Histograms of memorization scores for different CIFAR10 subsets}
    \label{fig:appendix_cifar10_separate}
\end{figure}

For completeness, we provide the histogram of memorization scores for different CIFAR10 subsets. The results are presented in \ref{fig:apendix_memscores}. This plot is obtained with the same data as the ones provided in the main part of the paper. Due to a different number of samples in each subset, bars have different heights, making the whole plot less readable and hard to interpret. For this reason, it was swapped for the mean memorization scores plot presented here on the right-hand side. We want to comment on the decrease of the mean for the subset of size 0.1. A general shift in the distribution of memorization scores probably causes it. In Fig.~\ref{fig:appendix_cifar10_separate} we present the histograms for different subset sizes on separate plots. For a subset of size 0.1, we can see that the whole distribution of memorization scores shifted its mode from around zero to 0.1. This caused the skew in the mean value and fraction of memorization scores above 0.25. We did not discuss this in the main part of our paper due to the limitation of page size.

\begin{figure}
    \centering
    \includegraphics[width=0.7\linewidth]{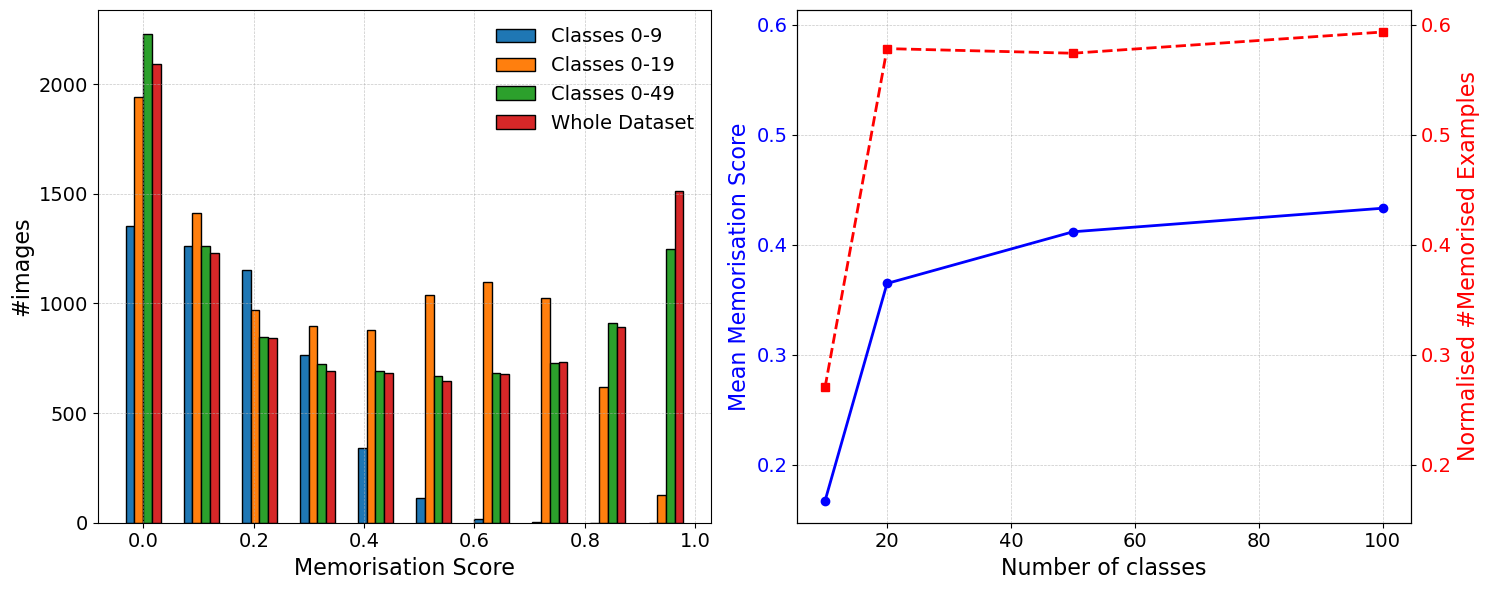}
    \includegraphics[width=0.7\linewidth]{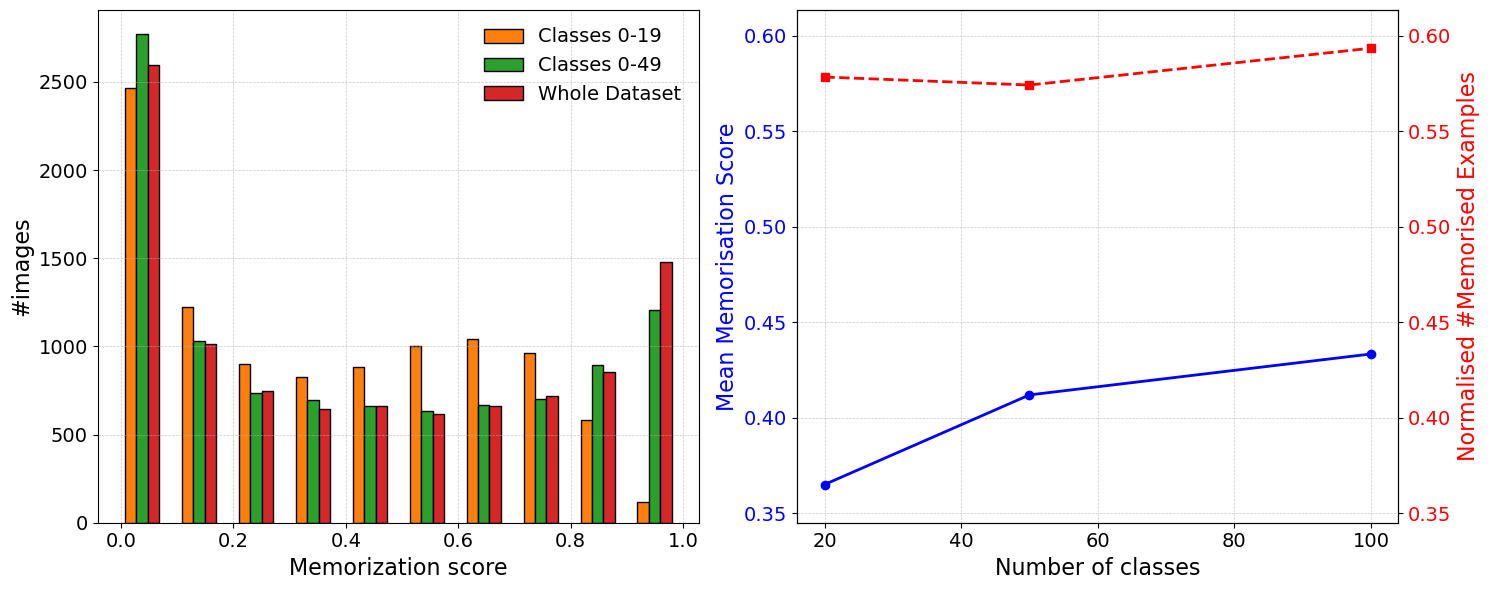}
    \includegraphics[width=0.7\linewidth]{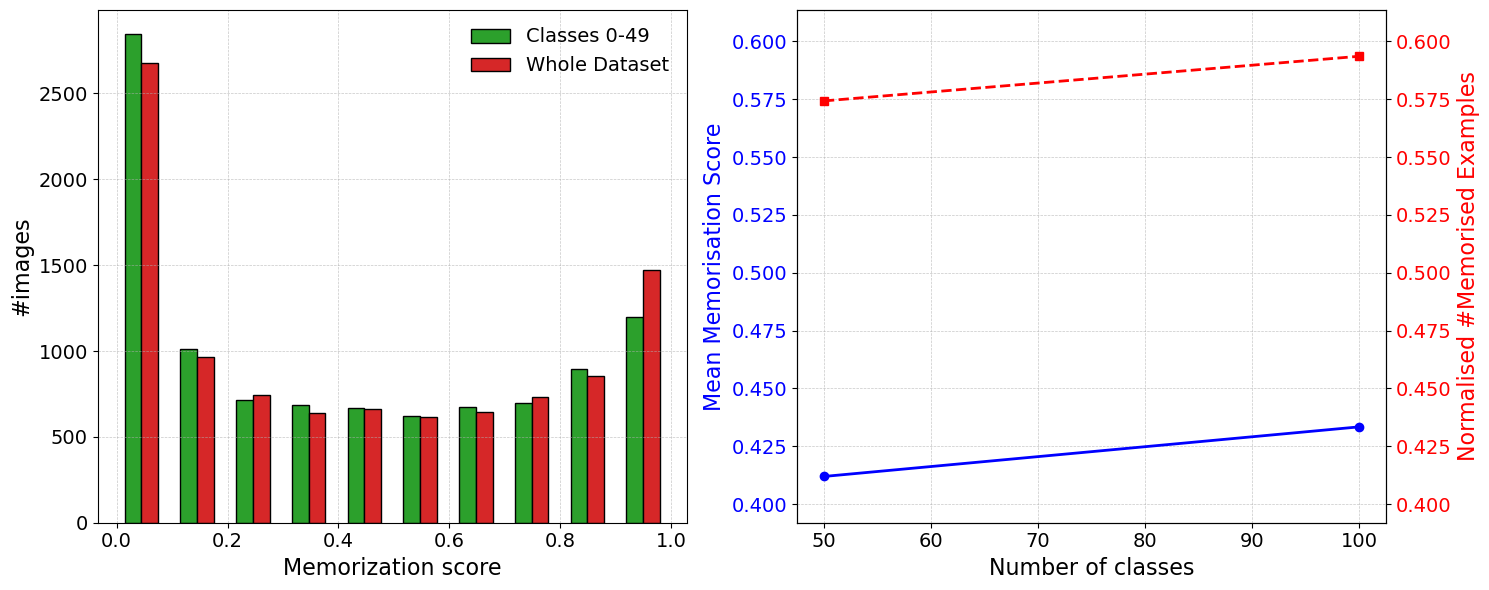}
    \caption{Histograms of memorization scores (left) and mean memorization score with normalized number of examples (right) for the first 10 (top), 20 (middle), and 50 (bottom) classes.}
    \label{fig:apendix_class_subsets_mem}
\end{figure}

The main part of our paper has memorization scores computed for the first 10 classes of CIFAR100 for different class subsets used for training. Here we provide plots for the first 20 and 50 classes as well in Fig.~\ref{fig:apendix_class_subsets_mem}.

In our earlier experiment description, we made the comment that the influence of the larger number of classes outweighs the influence of the smaller number of samples, without justifying it. 
With exact results for both CIFAR100 and CIFAR10 in this appendix, we can now justify our statement.
For 50 classes trained on CIFAR100 (Fig.~\ref{fig:apendix_class_subsets_mem} bottom), the mean memorization score is above 0.4. For CIFAR10 with 0.5 samples (see Fig.~\ref{fig:apendix_memscores} right), the mean memorization score is below 0.2. These networks were trained with an equal number of samples, but with different numbers of classes, showing that indeed the number of classes has a stronger influence than the number of samples.

\section{Memorization scores for Tiny ImageNet}
\label{app:tinyimagenet_memscores}

\begin{figure}
    \centering
    \includegraphics[width=0.8\linewidth]{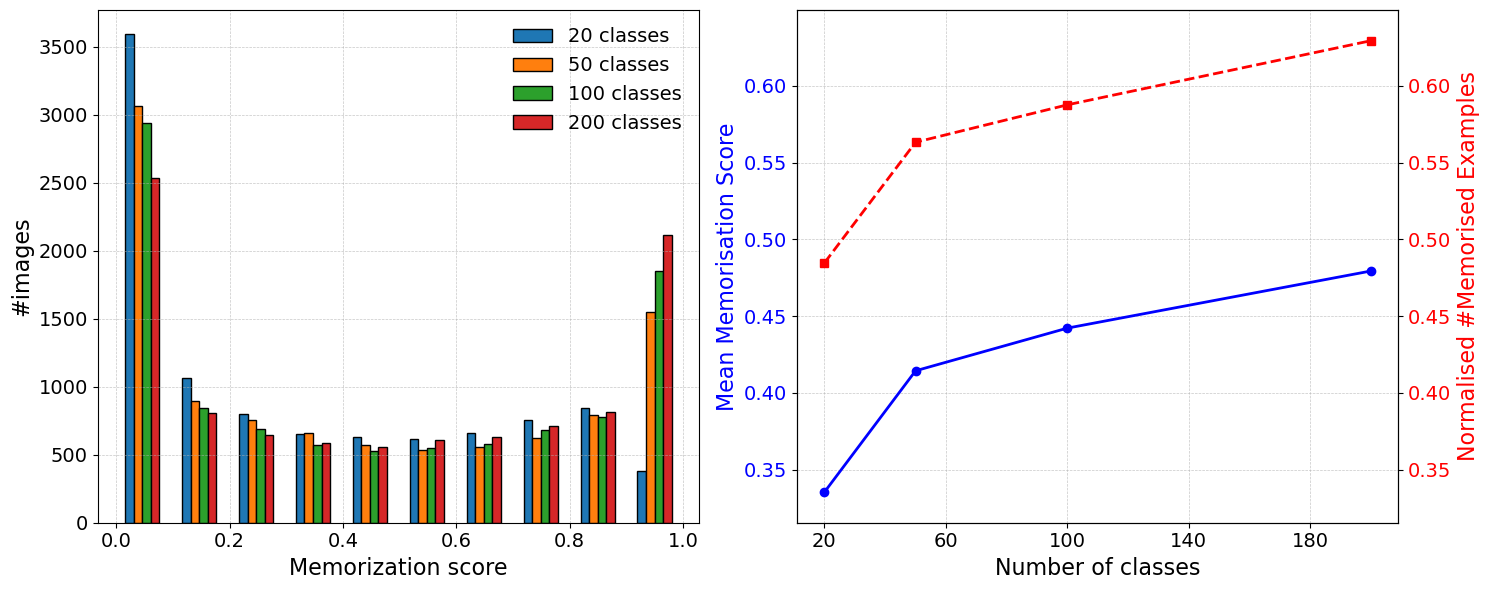}
    \caption{(Left) Histogram of proxy memorization scores for TinyImageNet dataset with various numbers of classes. (Right) Mean memorization scores and fraction of samples from dataset with memorization score above 0.25.}
    \label{fig:apendix_memscores_tinyimagenet}
\end{figure}

We test if our observations of the relation between the number of classes and memorization are robust and occur also in another dataset. We test Tiny ImageNet \cite{Wu2017TinyIC}, as it has a larger image resolution and a larger number of classes. We present our results in Fig.~\ref{fig:apendix_memscores_tinyimagenet}. We notice a very similar trend to the one observed in the main part of the paper - both the average memorization score and fraction of samples with memorization scores above 0.25 rise. This additional experiment results increase credibility of our conclusion, however, more experiments are needed to check if these observations hold for datasets with even larger image resolution, a larger number of images, different hyperparameters, and with different network architectures.

\section{Test set accuracy for different memorization thresholds}
\label{app:memorization_thresholds}

\begin{figure}
    \centering
    \includegraphics[width=1.0\linewidth]{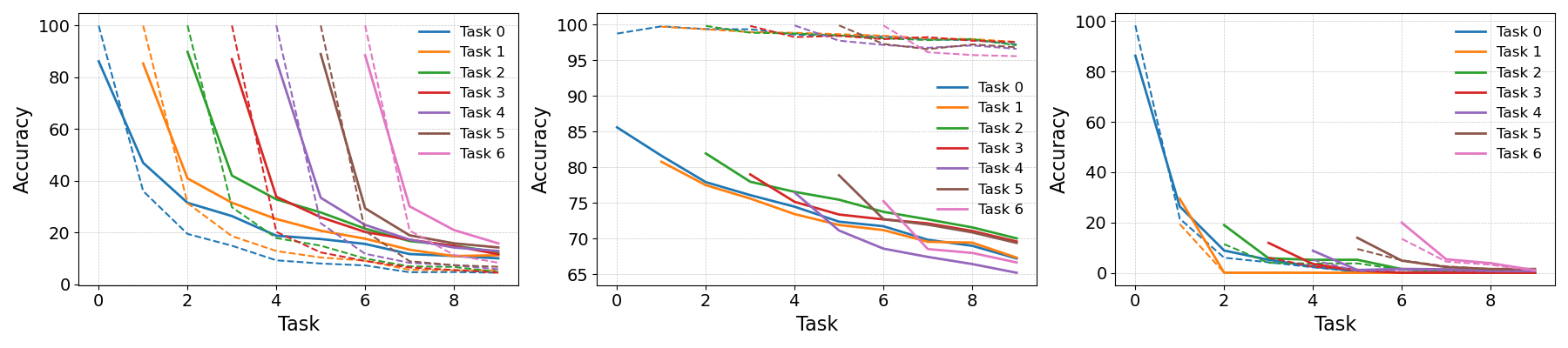}
    \includegraphics[width=1.0\linewidth]{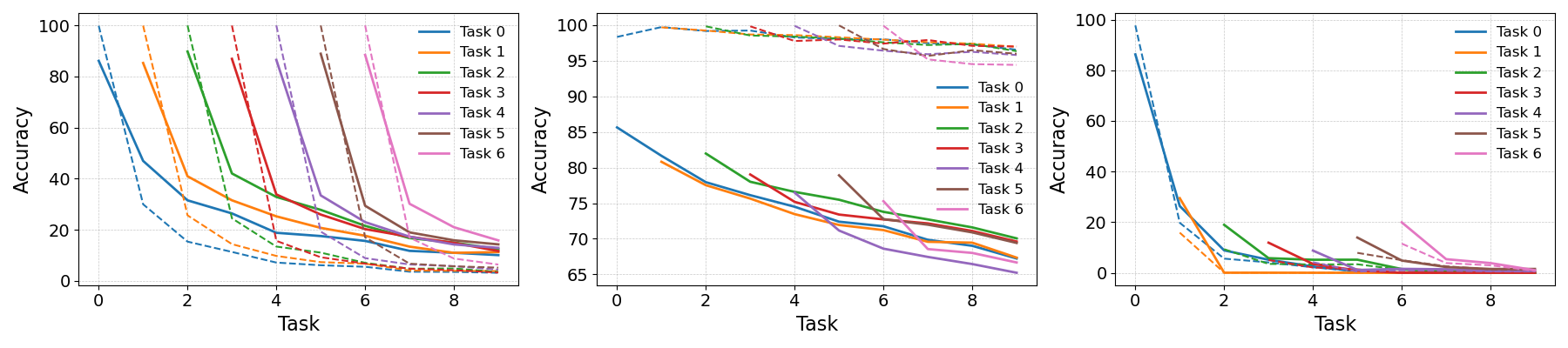}
    \includegraphics[width=1.0\linewidth]{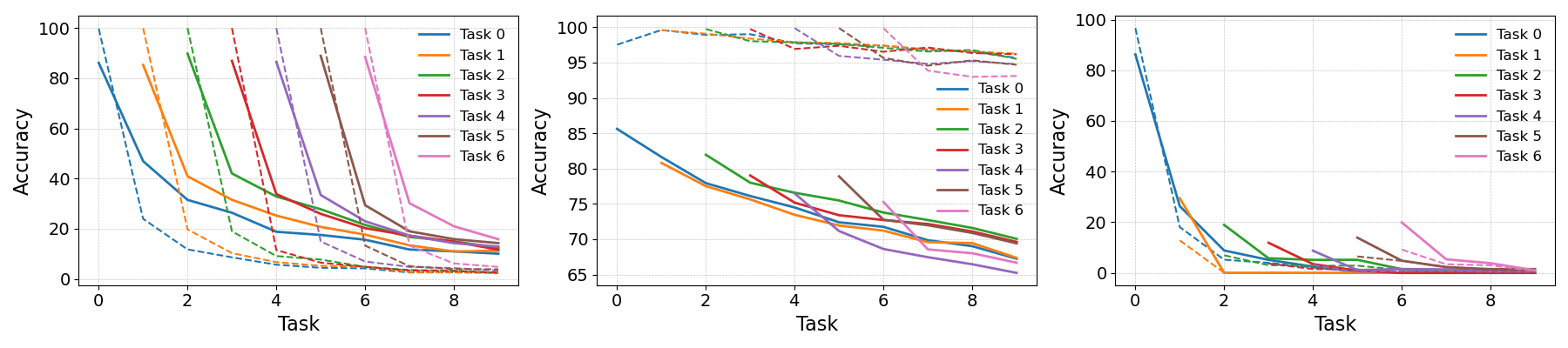}
    \includegraphics[width=1.0\linewidth]{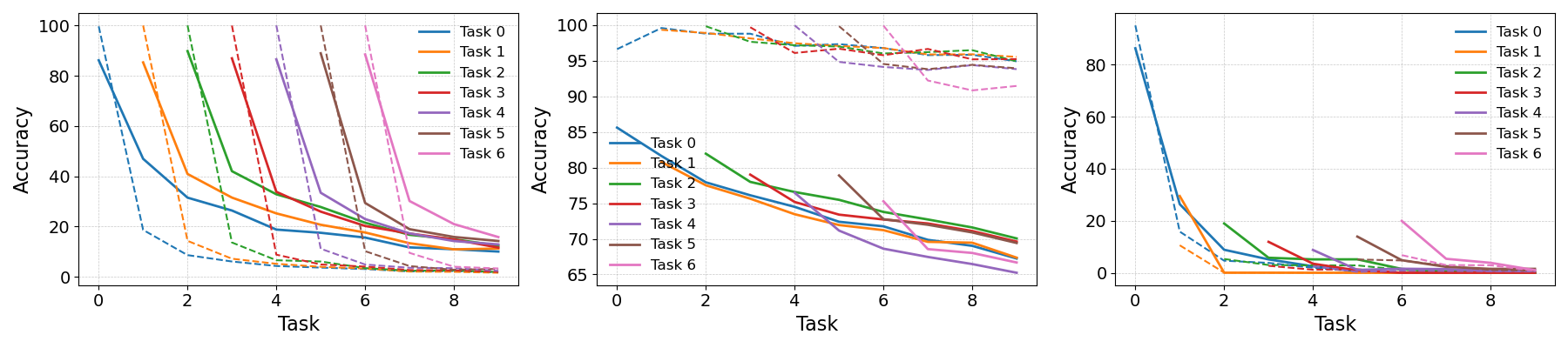}
    \caption{Task accuracy for test set (solid line) and long tail (dotted line) across incremental training on Seq-Cifar100 stream with 10 tasks. (Left) training with buffer size 500. (Middle) training with full access to previous tasks. (Right) training with LwF. The rows correspond to memorization score thresholds of 0.25, 0.5, 0.75, 0.9. Results averaged over 5 runs.}
    \label{fig:memorization-incremental-thresholds}
\end{figure}

We provide results for memorization thresholds different from those used in the main part of the paper. The value of 0.25 used in the main part of the paper is consistent with prior literature \cite{feldman2020neuralnetworksmemorizewhy}, but it is rather arbitrary. For this reason, we provide our results with thresholds for determining memorized samples equal to 0.5, 0.75, and 0.9. The results plotted in Fig.~\ref{fig:memorization-incremental-thresholds} suggest that the general trends we reported in the paper also hold for other memorization score thresholds.

\section{Implementation and full algorithm}
\label{app:full_algo}

We provide the full algorithm for Memorization-aware Experience Replay in Listing~\ref{alg:maer}.
In case of top-k selector, we do not consider samples that were never correctly classified (i.e., we ignore samples with $v_i = \infty$.
In Listing~\ref{alg:maer}, we use mathematical notation in line 8 to indicate obtaining the index of samples in the dataset (at what position in the dataset the given sample is stored). In practice, however, we modify the implementation of the dataset to return the index along the image and label pair, to avoid the expensive search for a matching sample. 
In line 23, we update the buffer by selecting the most frequent classes and replacing the buffer samples that we randomly draw from the current most frequent class. 


\begin{algorithm}
\caption{Memorization-aware Experience Replay}
\label{alg:maer}
\begin{algorithmic}[1]
\Require $S = \{ D_1, D_2,... \} $ - stream with tasks, $f(\theta)$ - network, $\mathcal{M}$ - memory buffer, $Q$ - selector (top-k, or other)
\State $t \gets 0$
\While {$D_t$ arrives} 
    \State $v \gets [\infty, \infty, ..., \infty]^T \in \mathbb{R}^{n_t} $
    \State $iter \gets 0$
    \For{$X,Y \sim D_t$}
        \State $\hat{Y} \gets f(X,\theta)$
        \For{$x, y, \hat{y} \in X,Y,\hat{Y}$}    
            \State $i \gets D_{index}[(x,y)]$  
            \If{$\hat{y} = y$ AND $v_i = \infty$}
                \State $v_i \gets iter$
            \EndIf
            \If{$\hat{y} \ne y$ AND $v_i \ne \infty$}
                \State $v_i \gets \infty$
            \EndIf
        \EndFor
        \State $X_m, Y_m \gets \mathcal{M}$
        \State $\mathcal{L} \gets \mathcal{L}(\hat{Y}, Y) + \mathcal{L}(f(X_m, \theta), Y_m)$
        \State $\theta \gets \theta - \lambda \nabla_{\theta} \mathcal{L} $
        \State \textit{reservoir\_sampling} ($\mathcal{M}, X, Y$)
        \State $iter \gets iter + 1$
    \EndFor

    \State $D_s \gets \emptyset$
    \For{$c \in C_t$}
        \State $k \gets \{ i : y_i = c, y_i \in D_t \}$
        \State $v_s \gets Q(v[k], |\mathcal{M}| / (t+1) )$
        \State $D_s \gets D_s \cup \{(x_j,y_j) \in D_t | j \in v_s \}$
    \EndFor
    \State update $\mathcal{M}$ with $D_s$
\EndWhile
\end{algorithmic}
\end{algorithm}

\section{Time Complexity}
\label{app:time_complexity}

We provide the relative execution time for the proposed method and other baselines for all datasets used in experiments in Tab.~\ref{tab:execution_time}. Many factors can impact the program execution time, such as specific CPU or GPU settings. We did not provide accurate benchmarks that consider these factors. We provide only rough estimates based on our logs. We do not claim that these values are correct execution time benchmarks, and they should not be treated as such. Another factor that could potentially impact the execution time is the implementation. We did not optimize for speed in any of the baselines that we used. This could also heavily impact the results of execution time benchmarks. 

\begin{table}
    \centering
    \caption{Relative execution time for the proposed method and baselines used in our experiments.}
    \begin{tabular}{cccc}
        \toprule
         \multirow{ 2}{*}{buffer policy} & \multicolumn{3}{c}{relative execution time} \\
         \cmidrule(r){2-4}
         & Split-Cifar10 & Split-Cifar100 & Split-TinyImageNet \\
         \midrule
         reservoir & 1.0 & 1.0 & 1.0  \\
         reservoir balanced & 0.98 & 1.01 & 1.00 \\ 
         rainbow memory & 1.06 & 1.10 & 1.03 \\  
         PBCS & 2.05 & 2.43 & 2.00 \\ 
         BCSR & 1.09 & 1.18 & 1.19 \\ 
         bottom-k memscores & 1.01 & 1.06 & 1.00 \\
         middle-k memsocre & 1.01 & 1.06 & 1.00 \\
         top-k memscores & 1.03 & 1.10 & 1.03 \\
         \bottomrule
    \end{tabular}
    \label{tab:execution_time}
\end{table}

\section{Computational resources}
\label{app:compute_resources}

We used three machines to perform computations in this work:
\begin{itemize}
    \item machine with: 1x NVIDIA RTX3090 and 128Gb RAM memory
    \item machine with 2x NVIDIA RTX3090 and 64Gb RAM memory
    \item server with 8x NVIDIA A5000 and 128 Gb RAM memory
\end{itemize}

Training a single ResNet18 on full Cifar100 required approximately 10 minutes for our implementation. We used the standard implementation of Pytorch training, as we needed flexibility in terms of creating subsets of the data in a controlled manner. 
We provide the GPU hours required to reproduce our experiments:
\begin{itemize}
    \item impact of number of classes on memorization Cifar100: 82 hours 
    \item impact of dataset size on memorization: 113 hours 
    \item impact of depth on memorization: 185 hours (only ResNet34 and ResNet50, as ResNet18 was part of the previous experiment) 
    \item impact of width on memorization: 147 hours (only widths 0.25, 0.5 and 0.75, full width was part of previous experiment) 
    \item impact of number of classes on memorization TinyImageNet: 290 hours 
    \item memorization in incremental training: 10 hours
    \item incremental representation learning: 5 hours
    \item memorization score proxy: 51 hours
    \item evaluation with standard benchmarks: 241 hours (only the proposed methods, not the baselines)   
    \item training with larger buffers: 148 hours (only the proposed methods, not the baselines)  
\end{itemize}
We round up the numbers above to a full hour. 
Some of the experiments were run on both RTX3090 and A5000 GPUs, therefore we do not differentiate between compute time on these different cards. The compute time should be valid for graphic cards with both compute capability and VRAM equal to or larger than ones provided by RTX3090. We do not provide the compute required for some of the experiments in the appendices, as it would be hard to obtain exact values.
We estimate the full duration of the experiments of our whole project to be 1476 hours. 
Full research project required larger compute resources than one stated above for the running of experiments, as we did several preliminary experiments, and some runs failed due to errors in code or configuration.





\newpage
\clearpage
\section*{NeurIPS Paper Checklist}

\begin{enumerate}

\item {\bf Claims}
    \item[] Question: Do the main claims made in the abstract and introduction accurately reflect the paper's contributions and scope?
    \item[] Answer: \answerYes{} 
    \item[] Justification: We belive that claims made in the abstract and introduction reflect well the content of the paper and result of our experiments. 
    \item[] Guidelines:
    \begin{itemize}
        \item The answer NA means that the abstract and introduction do not include the claims made in the paper.
        \item The abstract and/or introduction should clearly state the claims made, including the contributions made in the paper and important assumptions and limitations. A No or NA answer to this question will not be perceived well by the reviewers. 
        \item The claims made should match theoretical and experimental results, and reflect how much the results can be expected to generalize to other settings. 
        \item It is fine to include aspirational goals as motivation as long as it is clear that these goals are not attained by the paper. 
    \end{itemize}

\item {\bf Limitations}
    \item[] Question: Does the paper discuss the limitations of the work performed by the authors?
    \item[] Answer: \answerYes{} 
    \item[] Justification: We added separate section for discussing limitations of our findings. 
    \item[] Guidelines:
    \begin{itemize}
        \item The answer NA means that the paper has no limitation while the answer No means that the paper has limitations, but those are not discussed in the paper. 
        \item The authors are encouraged to create a separate "Limitations" section in their paper.
        \item The paper should point out any strong assumptions and how robust the results are to violations of these assumptions (e.g., independence assumptions, noiseless settings, model well-specification, asymptotic approximations only holding locally). The authors should reflect on how these assumptions might be violated in practice and what the implications would be.
        \item The authors should reflect on the scope of the claims made, e.g., if the approach was only tested on a few datasets or with a few runs. In general, empirical results often depend on implicit assumptions, which should be articulated.
        \item The authors should reflect on the factors that influence the performance of the approach. For example, a facial recognition algorithm may perform poorly when image resolution is low or images are taken in low lighting. Or a speech-to-text system might not be used reliably to provide closed captions for online lectures because it fails to handle technical jargon.
        \item The authors should discuss the computational efficiency of the proposed algorithms and how they scale with dataset size.
        \item If applicable, the authors should discuss possible limitations of their approach to address problems of privacy and fairness.
        \item While the authors might fear that complete honesty about limitations might be used by reviewers as grounds for rejection, a worse outcome might be that reviewers discover limitations that aren't acknowledged in the paper. The authors should use their best judgment and recognize that individual actions in favor of transparency play an important role in developing norms that preserve the integrity of the community. Reviewers will be specifically instructed to not penalize honesty concerning limitations.
    \end{itemize}

\item {\bf Theory assumptions and proofs}
    \item[] Question: For each theoretical result, does the paper provide the full set of assumptions and a complete (and correct) proof?
    \item[] Answer: \answerNA{} 
    \item[] Justification: the paper does not include theoretical results. 
    \item[] Guidelines:
    \begin{itemize}
        \item The answer NA means that the paper does not include theoretical results. 
        \item All the theorems, formulas, and proofs in the paper should be numbered and cross-referenced.
        \item All assumptions should be clearly stated or referenced in the statement of any theorems.
        \item The proofs can either appear in the main paper or the supplemental material, but if they appear in the supplemental material, the authors are encouraged to provide a short proof sketch to provide intuition. 
        \item Inversely, any informal proof provided in the core of the paper should be complemented by formal proofs provided in appendix or supplemental material.
        \item Theorems and Lemmas that the proof relies upon should be properly referenced. 
    \end{itemize}

    \item {\bf Experimental result reproducibility}
    \item[] Question: Does the paper fully disclose all the information needed to reproduce the main experimental results of the paper to the extent that it affects the main claims and/or conclusions of the paper (regardless of whether the code and data are provided or not)?
    \item[] Answer: \answerYes{} 
    \item[] Justification: We provide information about settings used in paper. We provide information about implementation of the proposed algorithm in Appendix~\ref{app:full_algo} with listing containing pseudocode. Detailed list of hyperparametrs values is provided in Appendix~\ref{app:hyperparameters} due to page limitation. We also provide the code repository with code for our experiments. Repository contains the spectification of enviornement used in experiments with all packages used and required versions.
    \item[] Guidelines:
    \begin{itemize}
        \item The answer NA means that the paper does not include experiments.
        \item If the paper includes experiments, a No answer to this question will not be perceived well by the reviewers: Making the paper reproducible is important, regardless of whether the code and data are provided or not.
        \item If the contribution is a dataset and/or model, the authors should describe the steps taken to make their results reproducible or verifiable. 
        \item Depending on the contribution, reproducibility can be accomplished in various ways. For example, if the contribution is a novel architecture, describing the architecture fully might suffice, or if the contribution is a specific model and empirical evaluation, it may be necessary to either make it possible for others to replicate the model with the same dataset, or provide access to the model. In general. releasing code and data is often one good way to accomplish this, but reproducibility can also be provided via detailed instructions for how to replicate the results, access to a hosted model (e.g., in the case of a large language model), releasing of a model checkpoint, or other means that are appropriate to the research performed.
        \item While NeurIPS does not require releasing code, the conference does require all submissions to provide some reasonable avenue for reproducibility, which may depend on the nature of the contribution. For example
        \begin{enumerate}
            \item If the contribution is primarily a new algorithm, the paper should make it clear how to reproduce that algorithm.
            \item If the contribution is primarily a new model architecture, the paper should describe the architecture clearly and fully.
            \item If the contribution is a new model (e.g., a large language model), then there should either be a way to access this model for reproducing the results or a way to reproduce the model (e.g., with an open-source dataset or instructions for how to construct the dataset).
            \item We recognize that reproducibility may be tricky in some cases, in which case authors are welcome to describe the particular way they provide for reproducibility. In the case of closed-source models, it may be that access to the model is limited in some way (e.g., to registered users), but it should be possible for other researchers to have some path to reproducing or verifying the results.
        \end{enumerate}
    \end{itemize}

\item {\bf Open access to data and code}
    \item[] Question: Does the paper provide open access to the data and code, with sufficient instructions to faithfully reproduce the main experimental results, as described in supplemental material?
    \item[] Answer: \answerYes{} 
    \item[] Justification: We provide code in the form of link to annonimized github repository. We will also provide .zip package with repository to our submission. We have reviewed code submission policy and we believe, that our work is in line with it.
    \item[] Guidelines:
    \begin{itemize}
        \item The answer NA means that paper does not include experiments requiring code.
        \item Please see the NeurIPS code and data submission guidelines (\url{https://nips.cc/public/guides/CodeSubmissionPolicy}) for more details.
        \item While we encourage the release of code and data, we understand that this might not be possible, so “No” is an acceptable answer. Papers cannot be rejected simply for not including code, unless this is central to the contribution (e.g., for a new open-source benchmark).
        \item The instructions should contain the exact command and environment needed to run to reproduce the results. See the NeurIPS code and data submission guidelines (\url{https://nips.cc/public/guides/CodeSubmissionPolicy}) for more details.
        \item The authors should provide instructions on data access and preparation, including how to access the raw data, preprocessed data, intermediate data, and generated data, etc.
        \item The authors should provide scripts to reproduce all experimental results for the new proposed method and baselines. If only a subset of experiments are reproducible, they should state which ones are omitted from the script and why.
        \item At submission time, to preserve anonymity, the authors should release anonymized versions (if applicable).
        \item Providing as much information as possible in supplemental material (appended to the paper) is recommended, but including URLs to data and code is permitted.
    \end{itemize}

\item {\bf Experimental setting/details}
    \item[] Question: Does the paper specify all the training and test details (e.g., data splits, hyperparameters, how they were chosen, type of optimizer, etc.) necessary to understand the results?
    \item[] Answer: \answerYes{} 
    \item[] Justification: We provide information about datasets and benchmarks used in the main part of the paper, while the detailed hyperparameters values are provided in Appendix~\ref{app:hyperparameters}. 
    \item[] Guidelines:
    \begin{itemize}
        \item The answer NA means that the paper does not include experiments.
        \item The experimental setting should be presented in the core of the paper to a level of detail that is necessary to appreciate the results and make sense of them.
        \item The full details can be provided either with the code, in appendix, or as supplemental material.
    \end{itemize}

\item {\bf Experiment statistical significance}
    \item[] Question: Does the paper report error bars suitably and correctly defined or other appropriate information about the statistical significance of the experiments?
    \item[] Answer: \answerYes{} 
    \item[] Justification: We provide error bars on plots and standard deviation in tables with results. We do not provide error bars for plots with with task accuracy for test data and samples with memorization score above some threshold. These plots contain a lot of lines, adding error bars would make them incompressible and hard to read. We tried adding error bars to these plots, and seeing the outcome we decided against that. The results in these plots were averaged over five runs with different random seed. We do provide the MLFlow logs with results for each run in our repository for that experiments, so an inquisitive reader can find this information.
    
    We also do not provide error bars for histograms and plots with values of memorization scores. Repeating such experiments would require substantial computational power, as each memorization score evaluation for single type of network or dataset requires training 250 neural networks. In the text however we do provide the upperbound of Feldman estimator variance according to information provided in \cite{feldman2020neuralnetworksmemorizewhy}.
    \item[] Guidelines:
    \begin{itemize}
        \item The answer NA means that the paper does not include experiments.
        \item The authors should answer "Yes" if the results are accompanied by error bars, confidence intervals, or statistical significance tests, at least for the experiments that support the main claims of the paper.
        \item The factors of variability that the error bars are capturing should be clearly stated (for example, train/test split, initialization, random drawing of some parameter, or overall run with given experimental conditions).
        \item The method for calculating the error bars should be explained (closed form formula, call to a library function, bootstrap, etc.)
        \item The assumptions made should be given (e.g., Normally distributed errors).
        \item It should be clear whether the error bar is the standard deviation or the standard error of the mean.
        \item It is OK to report 1-sigma error bars, but one should state it. The authors should preferably report a 2-sigma error bar than state that they have a 96\% CI, if the hypothesis of Normality of errors is not verified.
        \item For asymmetric distributions, the authors should be careful not to show in tables or figures symmetric error bars that would yield results that are out of range (e.g. negative error rates).
        \item If error bars are reported in tables or plots, The authors should explain in the text how they were calculated and reference the corresponding figures or tables in the text.
    \end{itemize}

\item {\bf Experiments compute resources}
    \item[] Question: For each experiment, does the paper provide sufficient information on the computer resources (type of compute workers, memory, time of execution) needed to reproduce the experiments?
    \item[] Answer: \answerYes{} 
    \item[] Justification: We provide that information in Appendix~\ref{app:compute_resources}. 
    \item[] Guidelines:
    \begin{itemize}
        \item The answer NA means that the paper does not include experiments.
        \item The paper should indicate the type of compute workers CPU or GPU, internal cluster, or cloud provider, including relevant memory and storage.
        \item The paper should provide the amount of compute required for each of the individual experimental runs as well as estimate the total compute. 
        \item The paper should disclose whether the full research project required more compute than the experiments reported in the paper (e.g., preliminary or failed experiments that didn't make it into the paper). 
    \end{itemize}
    
\item {\bf Code of ethics}
    \item[] Question: Does the research conducted in the paper conform, in every respect, with the NeurIPS Code of Ethics \url{https://neurips.cc/public/EthicsGuidelines}?
    \item[] Answer: \answerYes{} 
    \item[] Justification: We have carefully reviewed Code of Ethics and we believe that our work is complaint with it. 
    \item[] Guidelines:
    \begin{itemize}
        \item The answer NA means that the authors have not reviewed the NeurIPS Code of Ethics.
        \item If the authors answer No, they should explain the special circumstances that require a deviation from the Code of Ethics.
        \item The authors should make sure to preserve anonymity (e.g., if there is a special consideration due to laws or regulations in their jurisdiction).
    \end{itemize}

\item {\bf Broader impacts}
    \item[] Question: Does the paper discuss both potential positive societal impacts and negative societal impacts of the work performed?
    \item[] Answer: \answerYes{} 
    \item[] Justification: We provide broader impact discussion in appendix. 
    \item[] Guidelines:
    \begin{itemize}
        \item The answer NA means that there is no societal impact of the work performed.
        \item If the authors answer NA or No, they should explain why their work has no societal impact or why the paper does not address societal impact.
        \item Examples of negative societal impacts include potential malicious or unintended uses (e.g., disinformation, generating fake profiles, surveillance), fairness considerations (e.g., deployment of technologies that could make decisions that unfairly impact specific groups), privacy considerations, and security considerations.
        \item The conference expects that many papers will be foundational research and not tied to particular applications, let alone deployments. However, if there is a direct path to any negative applications, the authors should point it out. For example, it is legitimate to point out that an improvement in the quality of generative models could be used to generate deepfakes for disinformation. On the other hand, it is not needed to point out that a generic algorithm for optimizing neural networks could enable people to train models that generate Deepfakes faster.
        \item The authors should consider possible harms that could arise when the technology is being used as intended and functioning correctly, harms that could arise when the technology is being used as intended but gives incorrect results, and harms following from (intentional or unintentional) misuse of the technology.
        \item If there are negative societal impacts, the authors could also discuss possible mitigation strategies (e.g., gated release of models, providing defenses in addition to attacks, mechanisms for monitoring misuse, mechanisms to monitor how a system learns from feedback over time, improving the efficiency and accessibility of ML).
    \end{itemize}
    
\item {\bf Safeguards}
    \item[] Question: Does the paper describe safeguards that have been put in place for responsible release of data or models that have a high risk for misuse (e.g., pretrained language models, image generators, or scraped datasets)?
    \item[] Answer: \answerNA{} 
    \item[] Justification: Our work does not release any data or models. 
    \item[] Guidelines:
    \begin{itemize}
        \item The answer NA means that the paper poses no such risks.
        \item Released models that have a high risk for misuse or dual-use should be released with necessary safeguards to allow for controlled use of the model, for example by requiring that users adhere to usage guidelines or restrictions to access the model or implementing safety filters. 
        \item Datasets that have been scraped from the Internet could pose safety risks. The authors should describe how they avoided releasing unsafe images.
        \item We recognize that providing effective safeguards is challenging, and many papers do not require this, but we encourage authors to take this into account and make a best faith effort.
    \end{itemize}

\item {\bf Licenses for existing assets}
    \item[] Question: Are the creators or original owners of assets (e.g., code, data, models), used in the paper, properly credited and are the license and terms of use explicitly mentioned and properly respected?
    \item[] Answer: \answerYes{} 
    \item[] Justification: We credit libraries, tools and algortihms with proper citations. We provide versions of libraries used in environment specification in our repository. 
    \item[] Guidelines:
    \begin{itemize}
        \item The answer NA means that the paper does not use existing assets.
        \item The authors should cite the original paper that produced the code package or dataset.
        \item The authors should state which version of the asset is used and, if possible, include a URL.
        \item The name of the license (e.g., CC-BY 4.0) should be included for each asset.
        \item For scraped data from a particular source (e.g., website), the copyright and terms of service of that source should be provided.
        \item If assets are released, the license, copyright information, and terms of use in the package should be provided. For popular datasets, \url{paperswithcode.com/datasets} has curated licenses for some datasets. Their licensing guide can help determine the license of a dataset.
        \item For existing datasets that are re-packaged, both the original license and the license of the derived asset (if it has changed) should be provided.
        \item If this information is not available online, the authors are encouraged to reach out to the asset's creators.
    \end{itemize}

\item {\bf New assets}
    \item[] Question: Are new assets introduced in the paper well documented and is the documentation provided alongside the assets?
    \item[] Answer: \answerNA{} 
    \item[] Justification: We do not publish any assets in this paper. We only provide code with our experiments implementation, but we do not consider it an assets, as it not a library or framework. We believe that our code repository is well-documented. 
    \item[] Guidelines:
    \begin{itemize}
        \item The answer NA means that the paper does not release new assets.
        \item Researchers should communicate the details of the dataset/code/model as part of their submissions via structured templates. This includes details about training, license, limitations, etc. 
        \item The paper should discuss whether and how consent was obtained from people whose asset is used.
        \item At submission time, remember to anonymize your assets (if applicable). You can either create an anonymized URL or include an anonymized zip file.
    \end{itemize}

\item {\bf Crowdsourcing and research with human subjects}
    \item[] Question: For crowdsourcing experiments and research with human subjects, does the paper include the full text of instructions given to participants and screenshots, if applicable, as well as details about compensation (if any)? 
    \item[] Answer: \answerNA{} 
    \item[] Justification: Our work does not involve crowdsourcing nor research with human subjects. 
    \item[] Guidelines:
    \begin{itemize}
        \item The answer NA means that the paper does not involve crowdsourcing nor research with human subjects.
        \item Including this information in the supplemental material is fine, but if the main contribution of the paper involves human subjects, then as much detail as possible should be included in the main paper. 
        \item According to the NeurIPS Code of Ethics, workers involved in data collection, curation, or other labor should be paid at least the minimum wage in the country of the data collector. 
    \end{itemize}

\item {\bf Institutional review board (IRB) approvals or equivalent for research with human subjects}
    \item[] Question: Does the paper describe potential risks incurred by study participants, whether such risks were disclosed to the subjects, and whether Institutional Review Board (IRB) approvals (or an equivalent approval/review based on the requirements of your country or institution) were obtained?
    \item[] Answer: \answerNA{} 
    \item[] Justification: Our work does not involve crowdsourcing nor research with human subjects. 
    \item[] Guidelines:
    \begin{itemize}
        \item The answer NA means that the paper does not involve crowdsourcing nor research with human subjects.
        \item Depending on the country in which research is conducted, IRB approval (or equivalent) may be required for any human subjects research. If you obtained IRB approval, you should clearly state this in the paper. 
        \item We recognize that the procedures for this may vary significantly between institutions and locations, and we expect authors to adhere to the NeurIPS Code of Ethics and the guidelines for their institution. 
        \item For initial submissions, do not include any information that would break anonymity (if applicable), such as the institution conducting the review.
    \end{itemize}

\item {\bf Declaration of LLM usage}
    \item[] Question: Does the paper describe the usage of LLMs if it is an important, original, or non-standard component of the core methods in this research? Note that if the LLM is used only for writing, editing, or formatting purposes and does not impact the core methodology, scientific rigorousness, or originality of the research, declaration is not required.
    \item[] Answer: \answerNA{} 
    \item[] Justification: Our work does not involve LLMs. 
    \item[] Guidelines:
    \begin{itemize}
        \item The answer NA means that the core method development in this research does not involve LLMs as any important, original, or non-standard components.
        \item Please refer to our LLM policy (\url{https://neurips.cc/Conferences/2025/LLM}) for what should or should not be described.
    \end{itemize}

\end{enumerate}

\end{document}